\documentclass[a4paper, 11pt]{article}
\usepackage[left=3cm, right=3cm, top=3cm, bottom=3cm]{geometry}
\usepackage[utf8]{inputenc}
\usepackage{amsmath,amssymb,amsfonts}
\usepackage{authblk}
\makeatletter
\renewcommand\AB@affilsepx{~--- \protect\Affilfont}
\makeatother

\usepackage{libertine}
\usepackage[libertine]{newtxmath}
\usepackage[scaled=0.96]{zi4}

\usepackage[numbers]{natbib}
\title{Randomized Self Organizing Map}
\date{\small \today}
\author[1,2,3]{Nicolas P. Rougier}
\author[4]{Georgios Is. Detorakis}
\affil[1]{Inria Bordeaux Sud-Ouest}
\affil[2]{Institut des Maladies Neurodégénératives,
          Université  de Bordeaux, CNRS UMR 5293}
\affil[3]{LaBRI, Université de Bordeaux,
          Institut Polytechnique de Bordeaux, CNRS UMR 5800}
\affil[4]{adNomus Inc., San Jose, CA, USA}
\date{}

\usepackage[utf8]{inputenc}
\usepackage{amsmath,amssymb,amsfonts}
\usepackage{graphicx} 
\usepackage{lineno}
\usepackage{parskip}

\usepackage{xcolor}
\definecolor{bostonuniversityred}{rgb}{0.8, 0.0, 0.0}
\definecolor{blendedblue}{rgb}{0.2, 0.2, 0.6}
\definecolor{blendedred}{rgb}{0.8, 0.2, 0.2}
\definecolor{lightgray}{rgb}{0.5, 0.5, 0.5}
\usepackage[bookmarks=true,
         breaklinks=true,
         pdfborder={0 0 0},
         citecolor=blendedblue,
         colorlinks=true,
         linkcolor=blendedblue,
         urlcolor=blendedblue,
         citecolor=blendedblue,
         linktocpage=false,
         hyperindex=true,
         linkbordercolor=white]{hyperref}

\begin{document}
\maketitle
% Self-organizing maps usually rely on a predetermined topology of the neural space (the map), which is either a rectangular or a hexagonal Cartesian grid. When the intrinsic dimension of  the input space is much higher than the allowed dimension of the neural space, then the self-organizing map can be ill-formed. To overcome this problem in  high dimensional input spaces. We propose a variation of the self organizing map algorithm, where we consider random placement of neurons on a high-dimensional manifold. The positions of the neural units are drawn from a blue noise distribution from which various topologies can be derived. These topologies possess  random but controllable discontinuities that allow for a flexible self-organization, especially with high-dimensional data. The proposed algorithm has been tested on one-, two- and three-dimensions tasks as well as on MNIST handwritten digits dataset. Furthermore, we investigate the reorganization of the self-organizing maps when we either remove or add neurons to the map. To analyze the results we use spectral analysis and topological data analysis tools. 
%

\textbf{Abstract.} We propose a variation of the self organizing map algorithm by considering the random placement of neurons on a two-dimensional manifold, following a blue noise distribution from which various topologies can be derived. These topologies possess random (but controllable) discontinuities that allow for a more flexible self-organization, especially with high-dimensional data. The proposed algorithm is tested on one-, two- and three-dimensions tasks as well as on the MNIST handwritten digits dataset and validated using spectral analysis and topological data analysis tools. We also demonstrate the ability of the randomized self-organizing map to gracefully reorganize itself in case of neural lesion and/or neurogenesis.\par

% \textbf{Keywords.} Self Organization, Neural Networks, Vector Quantization, Voronoi Tesselation, Neural Map Topology

\setcounter{tocdepth}{2}
\tableofcontents
\pagebreak
\section{Introduction}

Self-organizing map \citep{Kohonen:1982} (SOM) is a vector quantization method that maps data onto a grid, usually two-dimensional and regular. After learning has converged, the codebook is self-organized such that the prototypes associated with two nearby nodes are similar. This is a direct consequence of the underlying topology of the map as well as the learning algorithm that, when presented with a new sample, modifies the code word of the best matching unit (BMU, the unit with the closest to the input code word) as well as the code word of units in its vicinity (neighborhood). SOMs have been used in a vast number of applications \citep{Kaski:1998,Oja:2003,Polla:2009} and today there exist several variants of the original algorithm \citep{Kohonen:2001}. However, according to the survey of \citep{Astudillo:2014}, only a few of these variants consider an alternative topology for the map, the regular Cartesian and the hexagonal grid being by far the most common used ones. Among the alternatives, the growing neural gas \citep{Fritzke:1994} is worth to be mentioned since it relies on a dynamic set of units and builds the topology {\em a posteriori} as it is also the case for the incremental grid growing neural network \citep{Blackmore:1995} and the controlled growth self organizing map \citep{Alahakoon:2000}. However, this {\em a posteriori} topology is built in the data space as opposed to the neural space. This means that the neighborhood property is lost and two neurons that are close to each other on the map may end with totally different prototypes in the data space. The impact of the network topology on the self-organization has also been studied by \citep{Jiang:2009} using the MNIST database. In the direct problem (evaluating influence of topology on performance), these authors consider SOMs whose neighborhood is defined by a regular, small world or random network and show a weak influence of the topology on the performance of the underlying model. In the inverse problem (searching for the best topology), authors try to optimize the topology of the network using evolutionary algorithms \citep{Eiben:2003} in order to minimize the classification error. Their results indicate a weak correlation between the topology and the performances in this specific case. However, \citep{Burguillo:2013} reported contradictory results to \citep{Eiben:2003}, when they studied the use of self-organizing map for time series predictions and considered different topologies (spatial, small-world, random and scale-free). They concluded that the classical spatial topology remains the best while the scale-free topology seems inadequate for the time series prediction task. But for the two others (random and small-world), the difference was not so large and topology does not seem to dramatically impact performance.

In this work, we are interested in exploring an alternative topology in order to specifically handle cases where the intrinsic dimension of the data is higher than the dimension of the map. 
Most of the time, the topology of the SOM is one dimensional (linear network) or two dimensional (regular or hexagonal grid) and this may not correspond to the intrinsic dimension of the data, especially in the high dimensional case. This may result in the non-preservation of the topology \citep{Villmann:1999} with potentially multiple foldings of the map. The problem is even harder considering the data are unknown at the time of construction of the network. To overcome this topological constraint, we propose a variation of the self organizing map algorithm by considering the random placement of neurons on a two-dimensional manifold, following a blue noise distribution from which various topologies can be derived. These topologies possess random discontinuities that allow for a more flexible self-organization, especially with high-dimensional data. After introducing the methods, the model will be illustrated and analyzed using several classical examples and its properties will be more finely introduced. Finally, we'll explain how this model can be made resilient to neural gain or loss by reorganizing the neural sheet using the centroidal Voronoi tesselation.

A constant issue with self-organizing maps is how can we measure the quality of a map. In SOM's literature, there is neither one measure to rule them all nor a single general recipe on how to measure the quality of the map. Some of the usual measures are the distortion \cite{rynkiewicz:2008}, the $\delta x - \delta y$ representation \citep{Demartines:1992}, and many other specialized measures for rectangular grids or specific types of SOMs \citep{Polani2002}. However, most of those measures cannot be used in this work since we do not use a standard grid for laying over the neural space, instead we use a randomly distributed graph (see supplementary material for standard measures). This and the fact that the neural space is discrete introduce a significant challenge on deciding what will be a good measure for our comparisons \citep{Polani2002} (i.e., to compare the neural spaces of RSOM and regular SOM with the input space). According to \citep{Polani2002}, the quality of the map's organization can be considered equivalent to topology preservation. Therefore, a topological tool such as the persistent homology can help in comparing the input space with the neural one. Topological Data Analysis (TDA) is a relatively new field of applied mathematics and offers a great deal of topological and geometrical tools to analyze point cloud data \citep{Carlsson:2009,HerculanoHouzel:2013}. Such TDA methods have been proposed in \citep{Polani2002}, however TDA wasn't that advanced and popular back then. Therefore, in this work we use the persistent homology and barcodes to analyze our results and compare the neural spaces generated by the SOM algorithms with the input spaces. We provide more details about TDA and persistent homology later in the corresponding section.

To avoid confusion between the original SOM proposed by Teuvo Kohonen and the newly randomized SOM, we'll refer to the original as \textbf{SOM} and the newly randomized one as \textbf{RSOM}.

\section{Methods}

\subsection{Notation}

In the following, we will use definitions and notations introduced by \citep{rougier:2011} where a neural map is defined as the projection from a manifold $\Omega \subset \mathbb{R}^d$ onto a set $\mathcal{N}$ of $n$ {\em  neuron}s which is formally written as $\Phi : \Omega \rightarrow \mathcal{N}$. Each neuron $i$ is associated with a code word $\mathbf{w}_i \in \mathbb{R}^d$, all of which establish the set  $\{\mathbf{w}_i\}_{i \in   \mathcal{N}}$ that is referred as the code book. The mapping from $\Omega$ to $\mathcal{N}$ is a closest-neighbor winner-take-all rule such that any vector $\mathbf{v} \in \Omega$ is mapped to a neuron $i$ with the code $\mathbf{w}_\mathbf{v}$ being closest to the actual presented stimulus vector $\mathbf{v}$,
\begin{equation}
\Phi : \mathbf{v} \mapsto argmin_{i \in \mathcal{N}} (\lVert \mathbf{v} -
\mathbf{w}_i \rVert).
\label{eq:psi}
\end{equation}
The neuron $\mathbf{w}_\mathbf{v}$ is named the best matching unit (BMU) and the set $C_i = \{x \in \Omega | \Phi(x) = \mathbf{w}_i \}$ defines the {\em receptive field} of the neuron $i$.

%Before we present our new SOM learning algorithm, we introduce the notation  and terminology we are using throughout the present work. We borrow the notation from a previous work \citep{rougier:2011}.  A neural map is defined to be the projection from a manifold $\Omega \subset \mathbb{R}^d$ onto a set $\mathcal{N}$ of $n$ {\em neuron}s $\Phi : \Omega \rightarrow \mathcal{N}$. Each neuron $i$ is associated with a code word $\mathbf{w}_i \in \mathbb{R}^d$, all of which establish the set  $\mathcal{W} = \{\mathbf{w}_i, i \in \mathcal{N}\}$ that is referred as the code book. The mapping from $\Omega$ to $\mathcal{N}$ is a closest-neighbor winner-take-all rule such that any vector $\mathbf{v} \in \Omega$ is mapped to a neuron $i$ with the code $\mathbf{w}_\mathbf{v}$ being closest to the current input vector $\mathbf{v}$,
%\begin{equation}
%\Phi : \mathbf{v} \mapsto argmin_{i \in \mathcal{N}} (\lVert \mathbf{v} -
%\mathbf{w}_i \rVert).
%\label{eq:psi}
%\end{equation}
%The neuron $\mathbf{w}_\mathbf{v}$ is named the best matching unit (BMU) and the set $C_i = \{x \in \Omega | \Phi(x) = \mathbf{w}_i \}$ defines the {\em receptive field} of neuron $i$.

\subsection{Spatial distribution} % \& Centroidal Voronoi Tesselation}
\label{sec:spatial_dist}

\begin{figure}[htbp]
  \includegraphics[width=\textwidth]{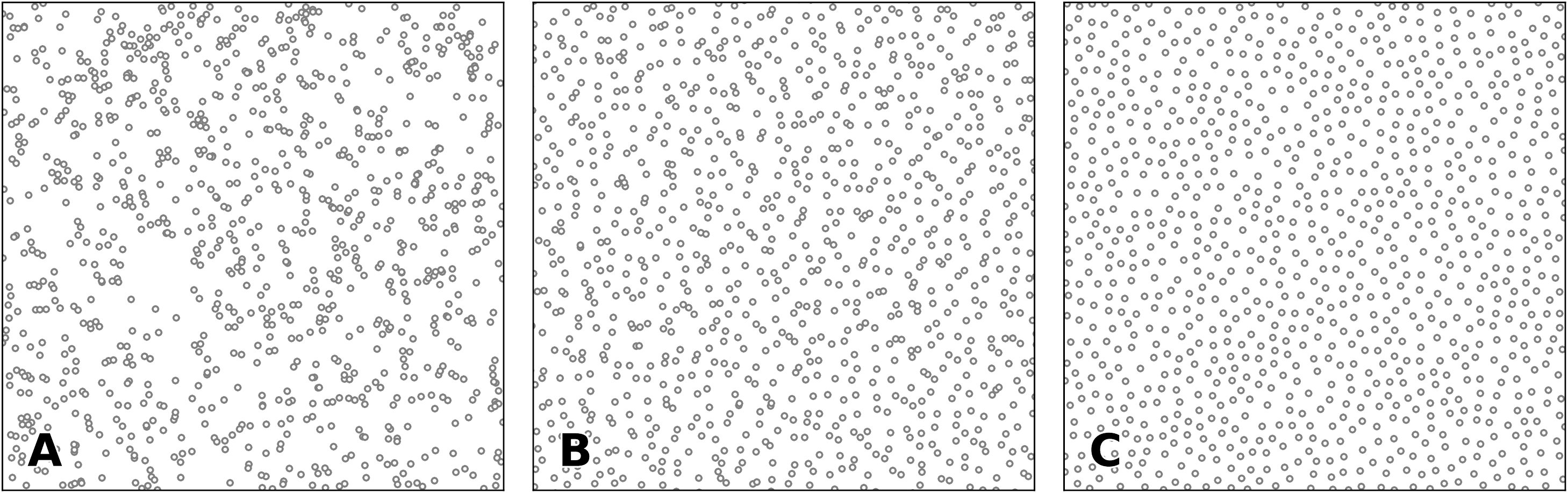}
  \caption{\textbf{Spatial distributions.}
    \textbf{\textsf{A.}} Uniform sampling (n=1000) corresponding to white noise.
    \textbf{\textsf{B.}} Regular grid (n=32$\times$32) + jitter (2.5\%).
    \textbf{\textsf{C.}} Poisson disc sampling (n=988) corresponding to blue noise.}
  \label{fig:sampling}
\end{figure}

The SOM space is usually defined as a two-dimensional region where nodes are arranged in a regular lattice (rectangular or hexagonal). Here, we consider instead the random placement of neurons with a specific spectral distribution (blue noise). As explained in \citep{Zhou:2012}, the spectral distribution property of noise patterns is often described in terms of the Fourier spectrum color. White noise corresponds to a flat spectrum with equal energy distributed in all frequency bands while blue noise has weak low-frequency energy, but strong high-frequency energy. In other words, blue noise has intuitively good properties with points evenly spread without visible structure (see figure~\ref{fig:sampling} for a comparison of spatial distributions).
There exists several methods \citep{Lagae:2008} to obtain blue noise sampling that have been originally designed for computer graphics (e.g. Poisson disk sampling, dart throwing, relaxation, tiling, etc.). Among these methods, the fast Poisson disk sampling in arbitrary dimensions \citep{Bridson:2007} is among the fastest ($\mathcal{O}(n)$) and easiest to use. This is the one we retained for the placement of neurons over the normalized region $[0,1]\times[0,1]$. Such Poisson disk sampling guarantees that samples are no closer to each other than a specified minimum radius. This initial placement is further refined by applying a LLoyd relaxation \citep{Lloyd:1982} scheme for 10 iterations, achieving a quasi centroidal Voronoi tesselation.

\subsection{Topology}
\label{sec:topo}

\begin{figure}
  \includegraphics[width=\columnwidth]{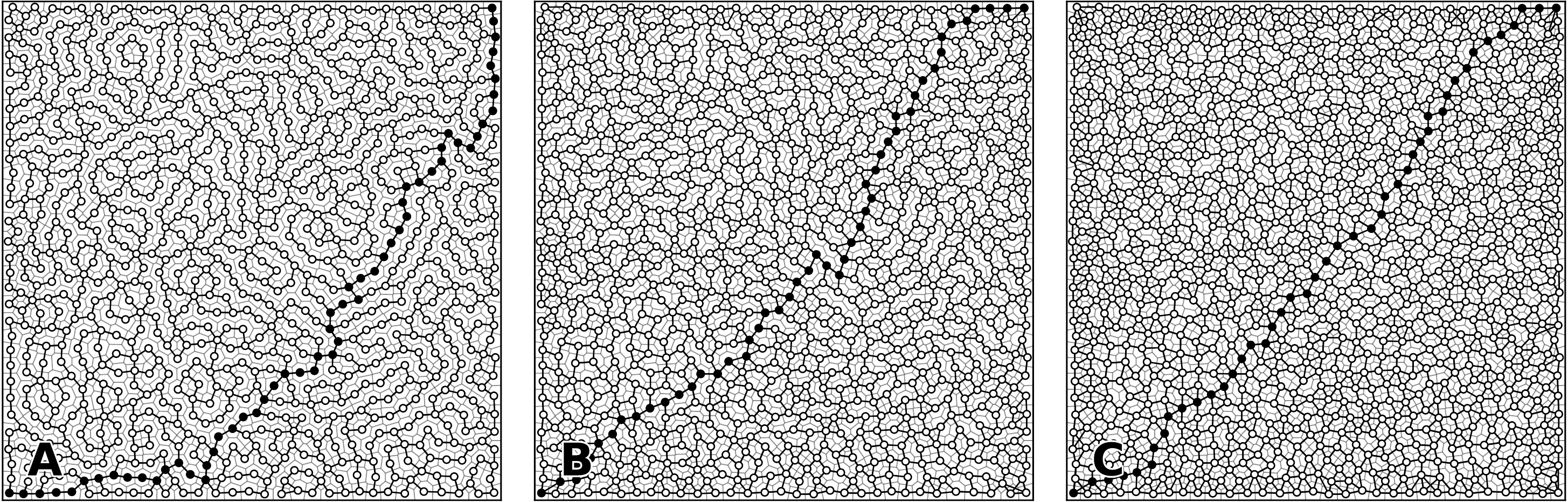}
  \caption{\textbf{Influence of the number of neighbours on the graph
    distance.} The same initial set of 1003 neurons has been equiped with
    2-nearest neighbors, 3 nearest neighbors and 4-nearest neighbors induced
    topology (panels \textbf{A}, \textbf{B} and \textbf{C} respectively). A
    sample path from the the lower-left neuron to the upper-right neuron has
    been highlighted with a thick line (with respective lengths of 59, 50 and
    46 nodes).}
  \label{fig:topology}
\end{figure}

Considering a set of $n$ points $P = \{P_i\}_{i \in [1,n]}$ on a finite region,
we first compute the Euclidean distance matrix $E$, where $e_{ij} = \lVert P_i - P_j \rVert$ 
and we subsequently define a connectivity matrix $G^{p}$
%= \{G^{p}_{ij}\}_{i,j \in [1,n]}$ \gid{$G^{p} = g^{p}_{ij},\, \text{where } i,j \in [1,n]$}
such that only the $p$ closest points
are connected. More precisely, if $P_j$ is among the $p$ closest neighbours of
$P_i$ then $g^p_{ij} = 1$ else we have $g^p_{ij} = 0$.
From this connectivity
matrix representing a graph, we compute the length of the shortest path between
each pair of nodes and stored them into a distance matrix $D^p$. Note that
lengths are measured in the number of nodes between two nodes such that two
nearby points (relatively to the Euclidean distance) may have a corresponding
long graph distance as illustrated in figure \ref{fig:topology}. This matrix
distance is then normalized by dividing it by the maximum distance between two
nodes such that the maximum distance in the matrix is 1. In the singular case
when two nodes cannot be connected through the graph, we recompute a spatial
distribution until all nodes can be connected.

\subsection{Learning}

The learning process is an iterative process between time $t=0$ and time $t=t_f \in \mathbb{N}^+$ where vectors $\mathbf{v} \in \Omega$ are sequentially presented to the map. For each presented vector $\mathbf{v}$ at time $t$, a winner $s \in \mathcal{N}$ is determined according to equation (\ref{eq:psi}). All codes $\mathbf{w}_{i}$ from the code book are shifted towards $\mathbf{v}$ according to
\begin{equation}
  \Delta\mathbf{w}_{i} = \varepsilon(t)~h_\sigma(t,i,s)~(\mathbf{v} -
  \mathbf{w}_i)
  \label{eq:som-learning}
\end{equation}
with $h_\sigma(t,i,j)$ being a neighborhood function of the form
\begin{equation}
  h_\sigma(t,i,j) = e^{- \frac{{d^p_{ij}}^2}{\sigma(t)^2}}
  \label{eq:som-neighborhood}
\end{equation}
where $\varepsilon(t) \in \mathbb{R}$ is the learning rate and $\sigma(t) \in \mathbb{R}$
is the width of the neighborhood defined as
\begin{equation}
  \sigma(t) =
  \sigma_i\left(\frac{\sigma_f}{\sigma_i}\right)^{t/t_f}, \text{ with } \varepsilon(t) =
  \varepsilon_i\left(\frac{\varepsilon_f}{\varepsilon_i}\right)^{t/t_f},
\end{equation}
while $\sigma_i$ and $\sigma_f$ are respectively the initial and final neighborhood width and $\varepsilon_i$ and $\varepsilon_f$ are respectively the initial and final learning rate. We usually have $\sigma_f \ll \sigma_i$ and $\varepsilon_f \ll \varepsilon_i$.

\subsection{Analysis Tools}
In order to analyze and compare the results of RSOM and SOM, we used a spectral method and persistence diagram analysis on the respective codebooks. These analysis tools are detailed below but roughly, the spectral method allows to estimate the distributions of eigenvalues in the activity of the maps while the persistence diagram allows to check for discrepancies between the topology of the input space and the topology of the map.

% To analyze the results of both the Kohonen SOM and VSOM algorithms and to make any comparison between the two algorithms we use a spectral method and persistence diagram on codebooks. The spectral method estimates the distributions of eigenvalues of the activity of neurons. The persistence diagram is a topological-geometrical  approach, more precisely is a tool coming from the field of topological data analysis (TDA). TDA provides the tools to investigate the topology of the maps and the input space and spot differences between the topology of the input space and the neural space of the SOM algorithms (Kohonen and VSOM). 

%\subsubsection{Topological Data Analysis}
\label{sec:tda}

Topological Data Analysis (TDA) \citep{Carlsson:2009} provides methods and tools to study topological structures of data sets such as point cloud and is useful when geometrical or topological information is not apparent within a data set. Furthermore, TDA tools are insensitive to dimension reduction and noise which make them well suited to analyze high-dimensional self-organized maps and their corresponding input data sets. In this work, we use the notion of persistent barcodes and diagrams \citep{Edelsbrunner:2008} to spot any differences between the topology of the input and neural spaces. Furthermore, we can apply some metrics from TDA such as the Bottleneck distance and measure how close two persistent diagrams are.
% qualify the quality of the representations of a map.
Since the exact manifold (or distribution) of the input space is not known in general and the SOM algorithms only approximate it, we simplify these manifolds by retaining their original topological structure.
Here we approach the manifolds of input and neural spaces using the Alpha complex. Before diving into more details regarding TDA, we provide here a few definitions and some notation. A $k$-simplex $\sigma$ is the convex hull of $k+1$ affinely independent points (for instance a $0$-simplex is a point, a $1$-simplex is an edge, a $2$-simplex is a triangle, etc). A simplicial complex with vertex set $\mathcal{V}$ is a set $\mathcal{S}$ of finite subsets of $\mathcal{V}$ such that the elements of $\mathcal{V}$ belong to $\mathcal{S}$ and for any $\sigma \in \mathcal{S}$ any subset $\sigma$ belongs to $\mathcal{S}$. Said differently, a simplicial complex is a space that has been  constructed out of intervals, triangles, and other higher dimensional simplices.

In our analysis we let $\mathcal{S}(\mathcal{M}, \alpha)$ be a Alpha simplicial complex with $\mathcal{M}$ being a point cloud, either the input space or the neural one, and $\alpha$ is the ``persistence'' parameter. More specifically, $\alpha$ is a threshold (or radius as we will see later) that determines if the set $X$ spans a $k$-simplex if and only if $d(x_i, x_j) \leq \alpha$ for all $0 \leq i, j \leq k$. From a practical point of view, we first define a family of thresholds $\alpha$ (or radius) and for each $\alpha$, we center a ball of radius $\alpha$ on each data point and look for possible intersections with other balls. This process is called filtration of simplicial complexes. We start from a small $\alpha$ where there are no intersecting balls (disjoint set of balls) and steadily we increase the size of $\alpha$ up to a point where a single connected blob emerges. As $\alpha$ varies from a low to a large value, holes open and close as different balls start intersecting. Every time an intersection emerges we assign a {\em birth} point $b_i$ and as the $\alpha$ increases and some new intersections of larger simplicies emerge some of the old simplicies die (since they merge with  other smaller simplicies to form larger ones). Then we assign a {\em death} point $d_i$. A pair of a birth and death points $(b_i, d_i)$ is plotted on a Cartesian two-dimensional plane and indicates when a simplicial complex was created and when it died. This two-dimensional diagram is called persistent diagram and the pairs (birth, death) that last longer reflect significant topological properties. The longevity of birth-death pairs is more clear in the persistent barcodes where the lifespan of such a pair is depicted as a straight line.

In other words, for each value of $\alpha$ we obtain new simplicial complexes and thus new topological properties such as  homology are revealed. Homology encodes the number of  points, holes, or voids in a space. For more thorough reading we refer the reader to \citep{Chazal:2017,Ghrist:2008,Zomorodian:2005}. In this work, we used the Gudhi library~\citep{Maria:2014} to compute the Alpha simplicial complexes, the filtrations and the persistent diagrams and barcodes. Therefore, we compute the persistent diagram and persistent barcode of the input space and of the maps and we calculate the Bottleneck distance between the input and SOM and RSOM maps diagrams. The bottleneck distance provides a tool to compare two persistent diagrams in a quantitative way. The Bottleneck distance between two persistent diagrams $\text{dgm}_1$ and $\text{dgm}_2$ as it is described in~\cite{Chazal:2017}
\begin{align}
    \label{eq:bottle}
    d_b(\text{dgm}_1, \text{dgm}_2) &= \inf_{\text{matching }m}\{ \max_{(p, q) \in m} \{||p - q||_{\infty} \} \},
\end{align}
where $p \in \text{dgm}_1 \backslash \Delta$, $q \in \text{dgm}_2 \backslash \Delta$, $\Delta$ is the diagonal of the persistent diagram (the diagonal $\Delta$ represents all the points that they die the very moment they get born, $b = d$). A matching between two  diagrams $\text{dgm}_1$ and $\text{dgm}_2$ is a subset $m \subset \text{dgm}_1 \times \text{dgm}_2$ such that every point in $\text{dgm}_1 \backslash \Delta$ and $\text{dgm}_2 \backslash \Delta$ appears exactly once in $m$. 

% In a similar way the Wasserstein distance is  defined by
% %%
% \begin{align}
%     \label{eq:wasser}
%     W_p(\text{dgm}_1, \text{dgm}_2)^p &= \inf_{\text{matching } m} \{ \sum_{(p, q) \in m}^{} ||p - q||^p_{\infty} \}.
% \end{align}
% %%

\subsection{Simulation Details}

Unless specified otherwise, all the models were parameterized using values given in table \ref{table:parameters}. These values were chosen to be simple and do not really impact the performance of the model. All simulations and figures were produced using the Python scientific stack, namely, SciPy \citep{Jones:2001}, Matplotlib \citep{Hunter:2007}, NumPy \citep{Walt:2011}, Scikit-Learn \citep{Pedregosa:2011}. Analysis were performed using Gudhi \citep{Maria:2014}). 
Sources are available at \href{https://github.com/rougier/VSOM}{github.com/rougier/VSOM}.
\begin{table}[!ht]
  \begin{center}
    \begin{tabular}{ll}
        \textbf{Parameter} & \textbf{Value} \\
        \hline
        Number of epochs      ($t_f$)           & 25000\\
        Learning rate initial ($\varepsilon_i$) & 0.50\\
        Learning rate final   ($\varepsilon_f$) & 0.01\\
        Sigma initial         ($\sigma_i$)      & 0.50\\
        Sigma final           ($\sigma_f$)      & 0.01\\
    \end{tabular}
      \caption{\textbf{Default parameters} Unless specified otherwise, these are
        the parameters used in all the simulations.}
      \label{table:parameters}
  \end{center}
\end{table}

\section{Results}

We ran several experiments to better characterize the properties of the randomized SOM and to compare them to the properties of a regular two-dimensional SOM. More specifically, we ran experiments using one dimensional, two dimensional and three dimensional datasets using uniform or shaped distributions. In this section, we only report a two-dimensional case and a three-dimensional case that we consider to be the most illustrative (all other results can be found in the supplementary material). We additionally ran an experiment using the MNIST hand-written data set and we compared it with a regular SOM. Finally, the last experiment is specific to the randomized SOM and shows how the model can recover from the removal (lesion) or the addition (neurogenesis) of neurons while conserving the overall self-organization. Each experiment (but the last) has been ran for both the randomized SOM and the regular SOM even though only the results for the randomized SOM are shown graphically in a dedicated figure while for the analysis, we use results from both SOM and RSOM. The reason to not show regular SOM results is that we assume the behavior is well known and does not need to be further detailed. 

\subsection{Two dimensional uniform dataset with holes}
\label{sec:2d-holes}

\begin{figure}
  \includegraphics[width=\columnwidth]{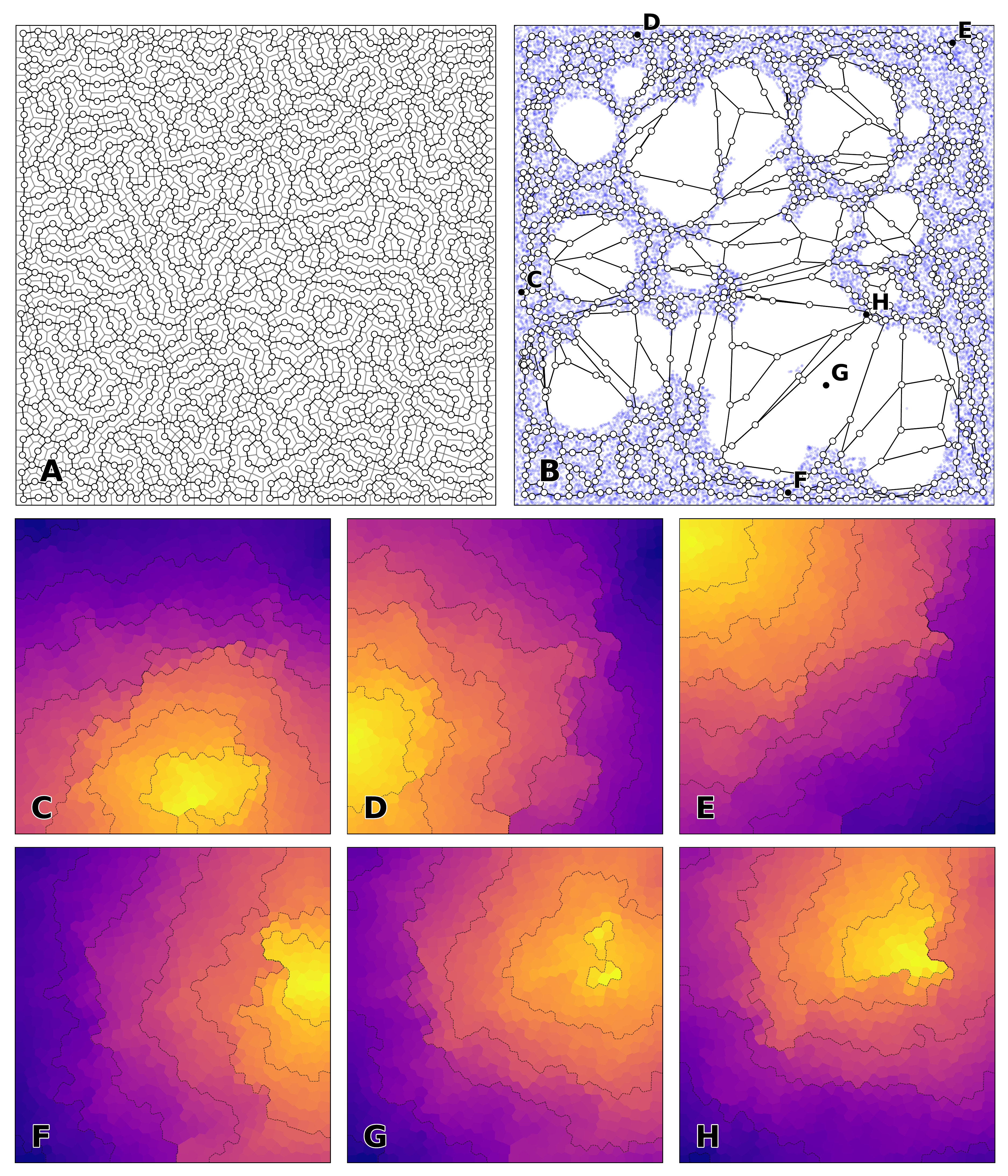}
  \vspace{2mm}
  \includegraphics[width=\columnwidth]{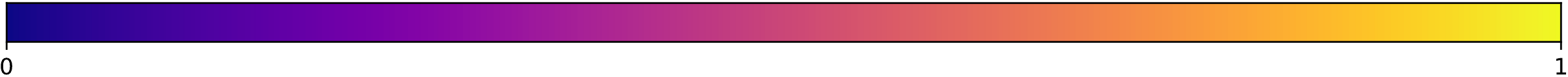}
  \caption{%
  {\bfseries \sffamily Two dimensional uniform dataset with holes (results)}
  Randomized SOM made of $1024$ neurons with a $2$-nearest neighbors induced topology. Model has been trained for $25,000$ epochs on two-dimensional points drawn from a uniform distribution on the unit square with holes of various sizes and random positions. \textbf{A} Map topology in neural space. \textbf{B} Map topology in data space. \textbf{C to H} Normalized distance map for six random samples. The \textbf{G} point has been purposely set outside the point distribution. Normalization has been performed for each sample in order to enhance contrast but this prevents comparison between maps.
  }
  \label{fig:2D-holes:results}
\end{figure}

In order to test for the adaptability of the randomized SOM to different topologies, we created a two dimensional uniform dataset with holes of various size and at random positions (see figure \ref{fig:2D-holes:results}B). Such holes are known to pose difficulties to the regular SOM since neurons whose codewords are over a hole (or in the immediate vicinity) are attracted by neurons outside the holes from all sides. Those neurons hence become dead units that never win the competition. In the RSOM, this problem exists but is less severe thanks to the absence of regularity in the underlying neural topology and the loose constraints (we use 2 neighbors to build the topology). This can be observed in \ref{fig:2D-holes:results}B) where the number of dead units is rather small and some holes are totally devoid of any neurons. Furthermore, when a sample that does not belong to the original distribution is presented, it can be observed that the answer of the map is maximal for a few neurons only (see figure \ref{fig:2D-holes:results}G)).

%The second experiment is designed to investigate how the SOM algorithms cope with two-dimensional uniformly distributed data points with holes ($x_1, x_2 \sim \mathcal{U}(0, 1)$). Therefore, in this case the input to the maps is two-dimensional Euclidean points on a plane, where we put some holes at random places on the plane (see Figure~\ref{Fig:persistence_exp2b} {\bfseries \sffamily B}, blue discs). We train again both the VSOM and the Kohonen maps over $25000$ samples for $25000$ epochs and each map consists of $1024$ neurons. For the VSOM we define the topology of the map using a blue noise distribution (see Figure~\ref{Fig:experiment2b} {\bfseries \sffamily A}) and for the Kohonen we use the standard rectangular Euclidean grid. After convergence both algorithms generate proper topographic maps covering the  input space. Figure~\ref{Fig:experiment2b} {\bfseries \sffamily B} shows the mapping of VSOM learning algorithm (white discs and black segments) along with the input space (blue discs). We observe that not too many neurons cover the holes.

\begin{figure}
  \centering
  \includegraphics[width=\columnwidth]{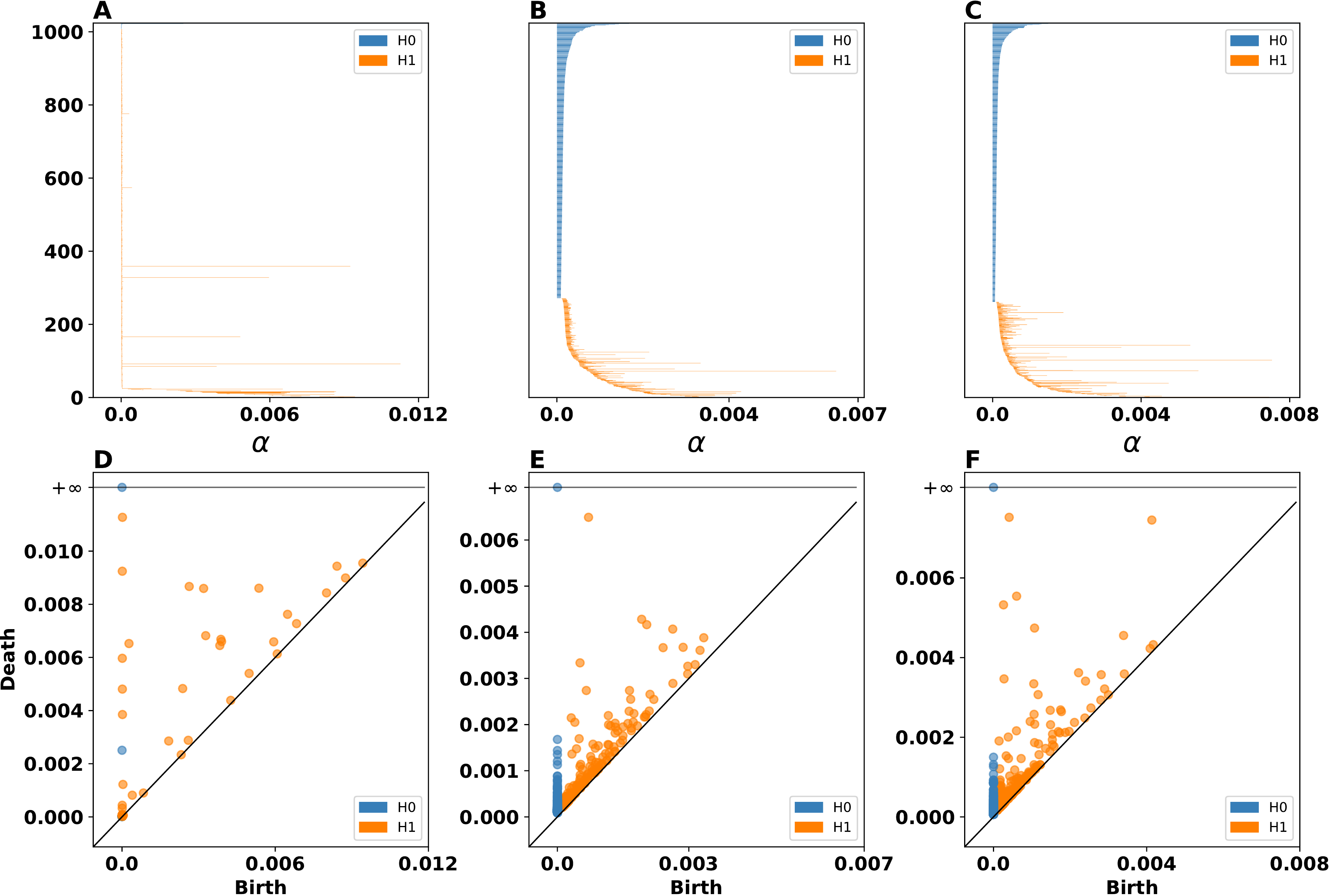}
  \caption{{\bfseries \sffamily Two dimensional uniform dataset with holes (analysis)}
  Persistent Barcodes of \textbf{A} input space, \textbf{B} SOM, and \textbf{RSOM}.
  The blue and orange line segments represent the $H0$- and $H1$-homology, respectively. This means
  that blue color represents connected segments within the space and orange color reflects the holes
  within the space. The longer the line segment the more important the corresponding topological
  feature. \textbf{D} illustrates the persistent diagram for the input space. \textbf{E} and \textbf{F}
  depict the persistent diagrams for SOM and RSOM, respectively. Again blue dots indicate $H0$-homology
  features and orange dots represent $H1$-homological features.}
  \label{fig:2D-holes:analysis}
\end{figure}

This observation is also supported by our topological analysis shown in figure \ref{fig:2D-holes:analysis}. Figures
\ref{fig:2D-holes:analysis}A, B, and C show the persistent barcodes where we can see the lifespan of each (birth, death)
pair (for more details about how we compute these diagrams see Section~\ref{sec:tda}). We observe that both the SOM and
the RSOM capture both the $H0$- and $H1$-homology of the input space, however the RSOM seems to have more persistent 
topological features for the $H1$-homology (orange lines). This means that the RSOM can capture more accurately the 
holes which are present in the input space. Roughly speaking we have about eight holes (see figure~\ref{fig:2D-holes:results}B) and we count about eight persistent features (the longest line segments
in the barcode diagrams) for RSOM in \ref{fig:2D-holes:analysis}C. On the other hand, the important persistent features
for the SOM are about five. In a similar way the persistent diagrams in figures~\ref{fig:2D-holes:analysis}C, D, and E
show that both RSOM and SOM capture in a similar way both the $H0$- and $H1$-homology features, although the 
RSOM (panel F) captures more holes as the isolated orange points away from the diagonal line indicate. This
is because the pairs that are further away from the diagonal are the most important meaning that they represent 
topological features that are the most persistent during the filtration process. 
Furthermore, we measure the Bottleneck distance between the persistence diagrams of input space and those of 
SOM and RSOM. The SOM's persistence diagram for $H0$ is closer to the input space (SOM: $0.000829$, RSOM: $0.001$),
while the RSOM's persistence diagram is closer to input's one for the $H1$ (SOM: $0.00478$, RSOM: $0.0037$). 
Finally, we ought to point out that the scale between panels A (D) and B, C (E, F) are not the same since the 
self-organization process has compressed information during mapping the input space to neural one.

\subsection{Three dimensional uniform dataset}

\begin{figure}
  \includegraphics[width=\columnwidth]{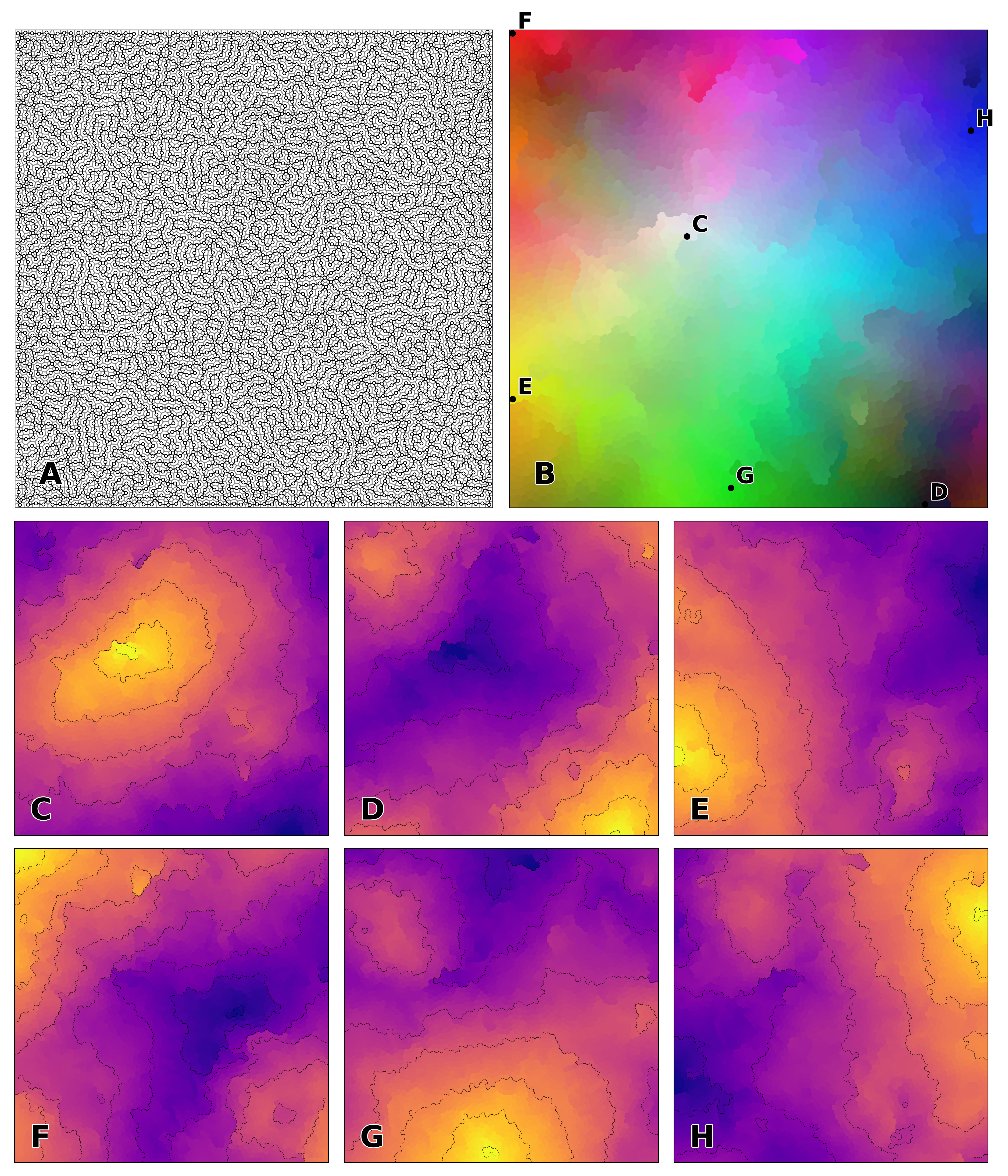}
  \vspace{2mm}
  \centering
  \includegraphics[width=.975\columnwidth]{colormap.pdf}
  \caption{%
  {\bfseries \sffamily Three dimensional uniform dataset (results)}
  Randomized SOM made of $4096$ neurons with a $3$-nearest neighbors induced topology. Model has been trained for $25,000$ epochs on three-dimensional points drawn from a uniform distribution on the unit cube. \textbf{A} Map topology in neural space. \textbf{B} Map codeword in neural space. Each neural voronoi cell is painted with the color of the codeword. \textbf{C to H} Normalized distance map for six samples, respectively (1,1,1), (0,0,0), (1,1,0), (1,0,0), (0,1,0) and (0,0,1) in RGB notations. Normalization has been performed for each sample in order to enhance contrast but this prevents comparison between maps.
  }
  \label{fig:3D-uniform:results}
\end{figure}

The three dimensional uniform dataset (that corresponds to the RGB color cube) is an interesting low dimensional case that requires a dimensionality reduction (from dimension 3 to 2). Since we used a uniform distribution this means the dataset is a dense three dimensional manifold that needs to be mapped to a two dimensional manifold which is known not to have an optimal solution. However, this difficulty can be partially alleviated using a loose topology in the RSOM. This is made possible by using a 2-neighbours induced topology as shown in figure \ref{fig:3D-uniform:results}A. This weak topology possesses several disconnected subgraphs that relax the constraints on the neighborhood of the BMU (see figure \ref{fig:topology-influence} in the supplementary section for the influence of neighborhood on the self-organization). This is clearly illustrated in figure \ref{fig:3D-uniform:results}B where the Voronoi cell of a neuron has been painted with the color of its codeword. We can observe an apparent structure of the RGB spectrum with some localized ruptures. To test for the completeness of the representation, we represented the position of six fundamental colors (C - white (1,1,1), D - black (0,0,0), E - yellow (1,1,0), F - red (1,0,0), G - green (0,1,0) and H - blue (0,0,1)) along with their associated distance maps after learning.

\begin{figure}
     \centering
     \includegraphics[width=\textwidth]{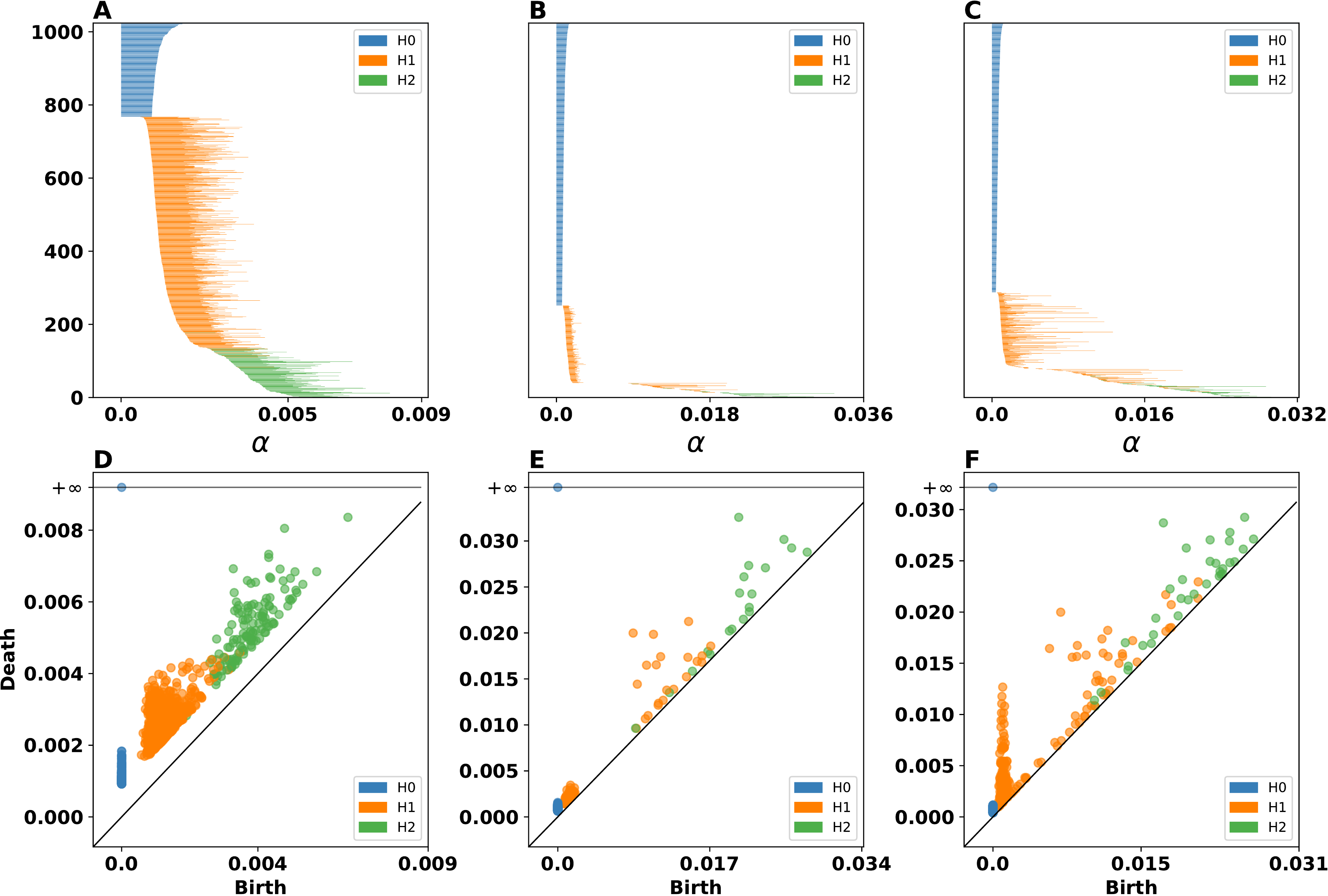}
     \caption{ {\bfseries \sffamily Three dimensional uniform dataset (analysis)}
     Persistent Barcodes of \textbf{A} input space, \textbf{B} SOM, and \textbf{RSOM}.
  The blue, orange, and green line segments represent the $H0$-, $H1$-, and $H2$-homology, respectively.
  This means that blue color represents connected segments within the space and orange color reflects the holes
  within the space and the green one the voids. The longer the line segment the more important the 
  corresponding topological feature. \textbf{D} illustrates the persistent diagram for the input space.
  \textbf{E} and \textbf{F} depict the persistent diagrams for SOM and RSOM, respectively. Again blue dots
  indicate $H0$-homology features, orange dots represent $H1$-homolocical features, and green the 
  $H2$-homological features.}
 \label{fig:3D-uniform:analysis}
\end{figure}

Furthermore, we performed the persistent homology to identify important topological features in the
input space and investigate how well the SOM and RSOM captured those features. Figures
\ref{fig:3D-uniform:analysis}A, B, and C show the persistent barcodes for the input space, SOM, and 
RSOM, respectively. We can see how the RSOM (panel C) captures more $H1$- and $H2$-homological properties
(since there are more persistent line segments, orange and green lines). The SOM (panel B) seems to capture
some of those features as well but they do not persist as long as they in the case of RSOM. The persistence 
diagrams of input, SOM and RSOM are shown in figures~\ref{fig:3D-uniform:analysis} D, E, and F, respectively.
These figures indicate that the RSOM has more persistent features (orange and green dots away from the diagonal
line) than the regular SOM.
The Bottleneck distance between the persistence diagrams of input space and those of SOM and RSOM reveals
that the SOM's persistence diagram is slightly closer to the input space's one for both the $H0$ (SOM: $0.00035$,
RSOM: $0.0007$), $H1$ (SOM: $0.006$, RSOM: $0.007$), and $H2$ (SOM: $0.0062$, RSOM: $0.0057$). Despite the 
fact that the bottleneck distances show that 
regular SOM's persistent diagram is closer to input space's one, the barcodes diagrams indicate that the RSOM 
captures more persistent topological features suggesting that RSOM preserves in a better way the topology of the
input space. Furthermore, the RSOM seems to capture better the higher dimensional topological features since the 
Bottleneck distances of $H2$-homological features are smaller for the RSOM than for the SOM.

% For the three-dimensional experiment we draw 50,000 three-dimensional points from a uniform distribution $\mathcal{U}([0, 1]\times [0, 1]\times [0, 1])$ (the cloud points form a cube in $\mathbb{R}^3$). The map is a two-dimensional manifold and thus the required task it to map the three-dimensional vectors  to the two-dimensional neural space. Once again we place the neurons on the two-dimensional neural space based on the sampling of a blue noise distribution and the induced topology is illustrated in Figure~\ref{fig:3D-uniform:results}A. The learning algorithm runs for $25000$ epochs over the three-dimensional vectors and after convergence we  obtain the map with the prototypes shown in Figure~\ref{fig:3D-uniform:results}B. Each color represents a different face of the three-dimensional cube (a cube has six faces so we obtain mainly six colors) and we observe that the SOM has mapped the three-dimensional cube on a two-dimensional neural space. The continuity and grouping of the colors indicates that the receptive fields of the neurons have been established properly. We illustrate this phenomenon in panels~\ref{fig:3D-uniform:results}C-H,  where the response of six different neurons (see the annotation in panel~\ref{fig:3D-uniform:results}B). The dark blue color indicates values close to zero and the yellow color represents values close to one.

\subsection{MNIST dataset}

\begin{figure}
  \includegraphics[width=\columnwidth]{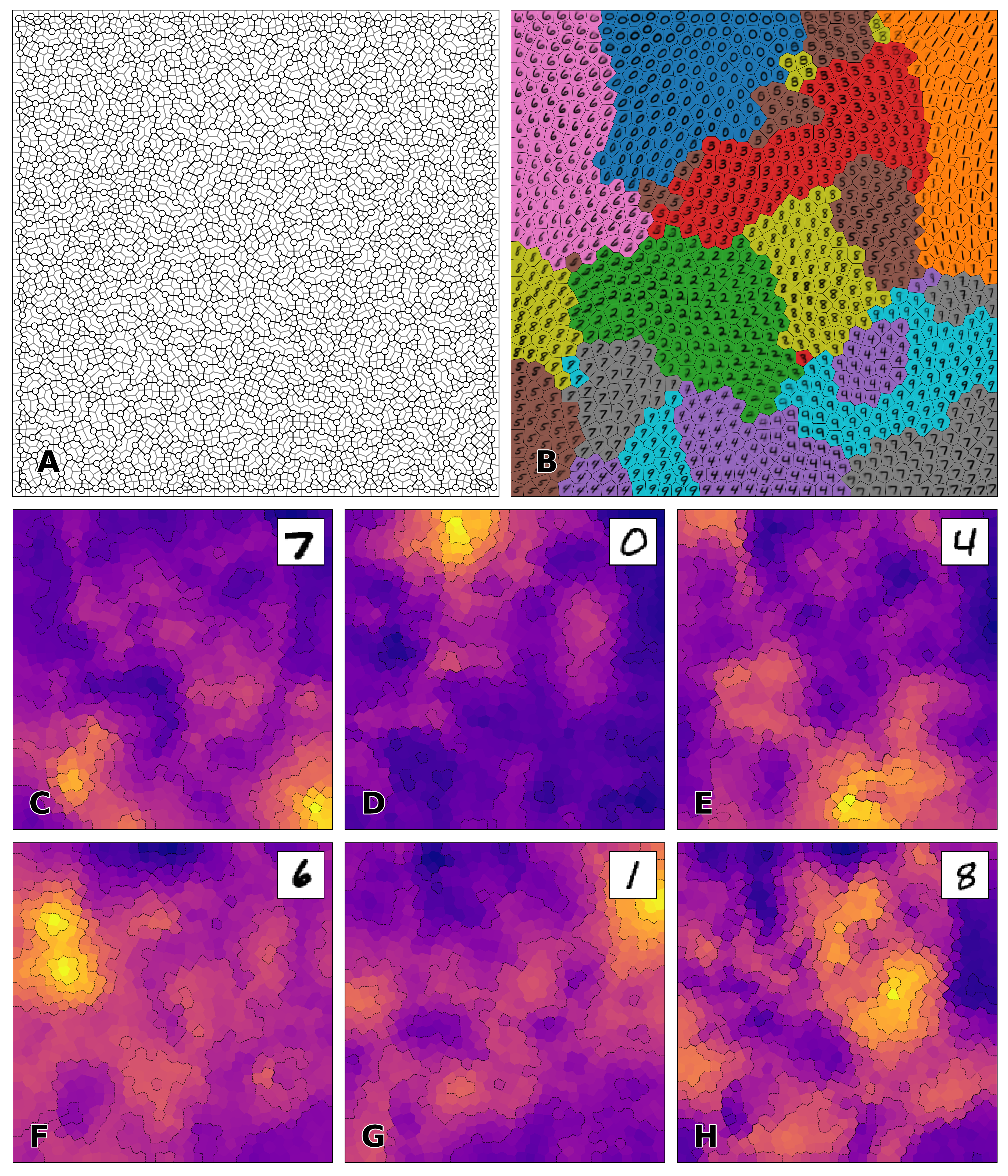}
  \vspace{2mm}
  \centering
  \includegraphics[width=.975\columnwidth]{colormap.pdf}
  \caption{%
  {\bfseries \sffamily MNIST dataset (results)}
  Randomized SOM made of $1024$ neurons with a $3$-nearest neighbors induced topology. Model has been trained for $25,000$ epochs on the MNIST dataset. \textbf{A} Map topology in neural space. \textbf{B} Map topology in data space. \textbf{C to H} Normalized distance map for six samples. Normalization has been performed for each sample in order to enhance contrast but this prevents comparison between maps.
  }
  \label{fig:MNIST:results}
\end{figure}

We tested RSOM on the standard MNIST dataset \citep{Lecun:1998} that contains $60,000$ training images and $10,000$ testing images. The dimension of each image is 28$\times$28 pixels and they are encoded using grayscale levels as the result of the normalization of the original black and white NIST database. The standard performance on most algorithms on the MNIST dataset is below 1\% error rate (with or without preprocessing) while for the regular SOM it is around 90\% recognition rate depending on the initial size, learning rate, and neighborhood function. Our goal here is not to find the best set of hyper-parameters but rather to explore if SOM and RSOM are comparable for a given set of hyper-parameters. Consequently, we considered a size of 32$\times$32 neurons and used the entire training set (60,000 examples) for learning and we measured performance on the entire testing set. We did not use any preprocessing stage on the image and we fed directly each image of the training set with the associated label to the model. Labels (from 0 to 9) have been transformed to a binary vector of size 10 using one-hot encoding (e.g. label 3 has been transformed to 0000001000). These binary labels can then be learned using the same procedure as for the actual sample. To decode the label associated to a code word, we simply consider the argmax of these binary vectors. Figure \ref{fig:MNIST:results} shows the final self-organisation of the RSOM where the class for each cell has been colorized using random colors. We can observe a number of large clusters of cells representing the same class (0, 1, 2, 3, 6) while the other classes (4,5,7,8,9) are split in two or three clusters. Interestingly enough, the codewords at the borders between two clusters are very similar. In term of recognition, this specific RSOM has an error rate just below 10\% ($0.903,\, \pm 0.296$) which is quite equivalent to the regular SOM error rate ($0.906,\, \pm 0.292$). The perfomances of the RSOM and SOM are actually not significantly different, suggesting that the regular grid hypothesis can be weaken.

\begin{figure}
  \centering
  \includegraphics[width=\textwidth]{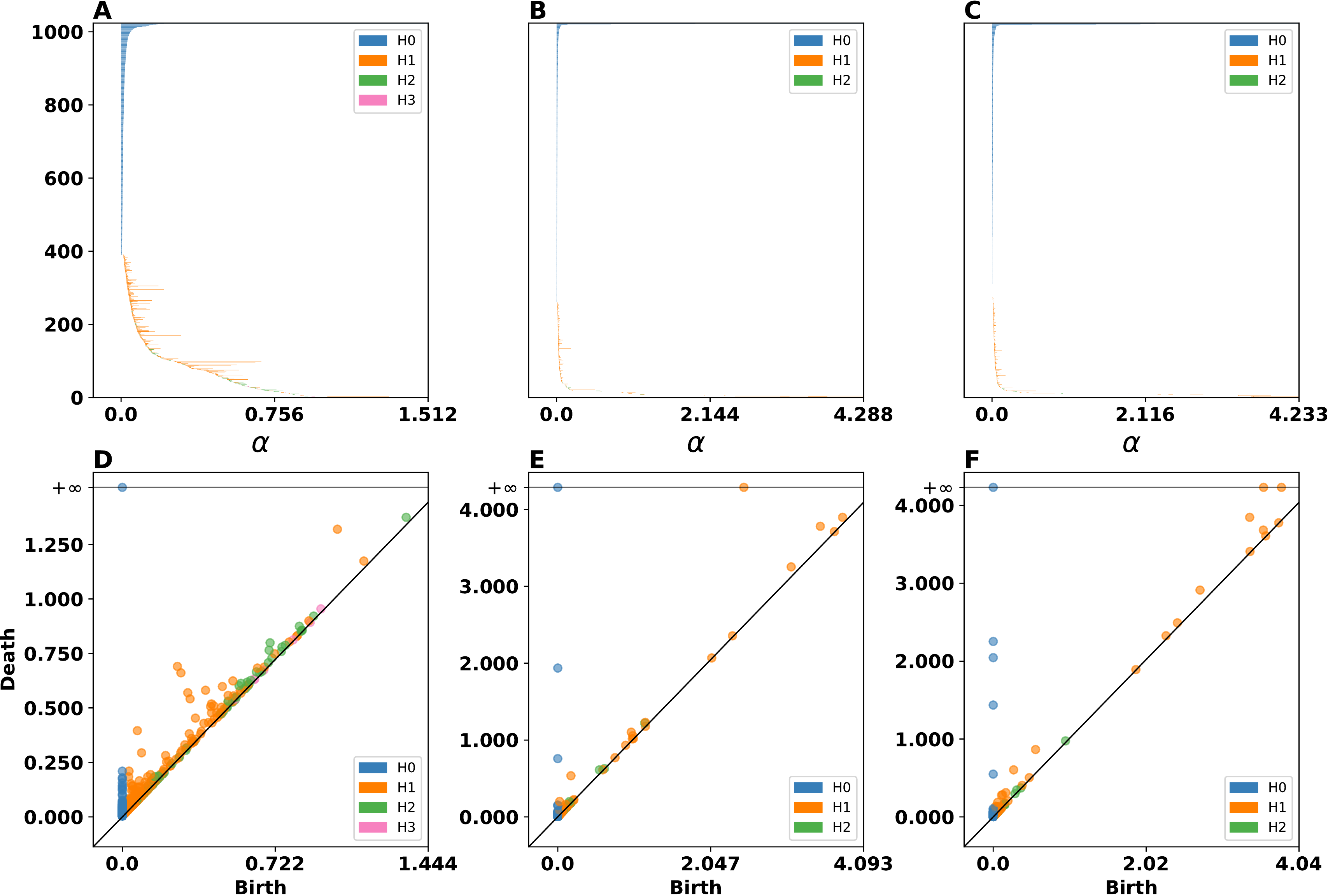}
  \caption{{\bfseries \sffamily MNIST dataset (analysis)}
  Persistent Barcodes of \textbf{A} input space, \textbf{B} SOM, and \textbf{RSOM}.
  The blue, orange, and green line segments represent the $H0$-, $H1$-, and $H2$-homology,
  respectively. This means
  that blue color represents connected segments within the space, orange color reflects the holes
  within the space and green the voids. The longer the line segment the more important the 
  corresponding topological feature. \textbf{D} illustrates the persistent diagram for the input space.
  \textbf{E} and \textbf{F} depict the persistent diagrams for SOM and RSOM, respectively. Again blue dots
  indicate $H0$-homology features, orange dots represent $H1$-homolocical features, and green the 
  $H2$-homological features.}
  \label{fig:MNIST:analysis}
\end{figure}

In a similar way we measured the similarity of the neural spaces generated by both the regular SOM and the RSOM using the persistent diagram and barcodes. The only significant difference from previous analysis was the projection of the input and neural spaces to a lower-dimension space via UMAP \citep{Mcinnes:2018}. Projections of high-dimensional spaces to lower-dimension ones have been used before in the analysis of latent spaces of autoencoders \citep{Detorakis:2019}. Here, we use the UMAP since it's an efficient and robust method for applying a dimensionality reduction on input and neural spaces. More precisely, we project the MNIST digits as well as the code words (dimension $784$) to a space of dimension $7$. Once we get the projections, we proceed to the topological analysis using the persistent diagram and barcodes as we already have described in previous paragraphs.
Figure~\ref{fig:MNIST:analysis} shows the results regarding the persistent barcodes and diagrams. 
The persistent barcodes in figures~\ref{fig:MNIST:analysis}A, B, and C indicate that RSOM captures more 
persistent features (panel C, orange and green lines reflect the $H1$- and $H2$-homological features, respectively)
than the regular SOM (panel B). The persistence diagrams of input, SOM and RSOM are shown in 
figures~\ref{fig:MNIST:analysis} D, E, and F, respectively. 
These figures indicate that the RSOM has more persistent features (orange and green dots away from the diagonal
line) than the regular SOM, consistently with the two previous experiments ($2$D uniform distribution with holes
and $3$D uniform distribution). The Bottleneck distance between the persistence diagrams of input space and 
those of SOM and RSOM for the $H0$ are SOM: $1.0$ and RSOM: $1.12$, for $H1$ SOM: $0.19$ and RSOM: $0.22$, and
finally for the $H2$ are SOM: $0.05$ and RSOM:$0.05$. Again we observe that the regular SOM has a persistent 
diagram that is closer to the one of the input space than that of RSOM, however the RSOM seems to approache 
slightly better the input space topology since it has more pairs (birth, death) away from the diagonal (black
line) in figures~\ref{fig:MNIST:analysis}D, E, and F. Moreover, the persistent barcode of RSOM (figure~\ref{fig:MNIST:analysis}C indicates that has more persistent features for the radius $\alpha$
between $0$ and $1.512$ than the regular SOM.

\subsection{Reorganization following removal or addition of neurons}

\begin{figure}
  \includegraphics[width=\columnwidth]{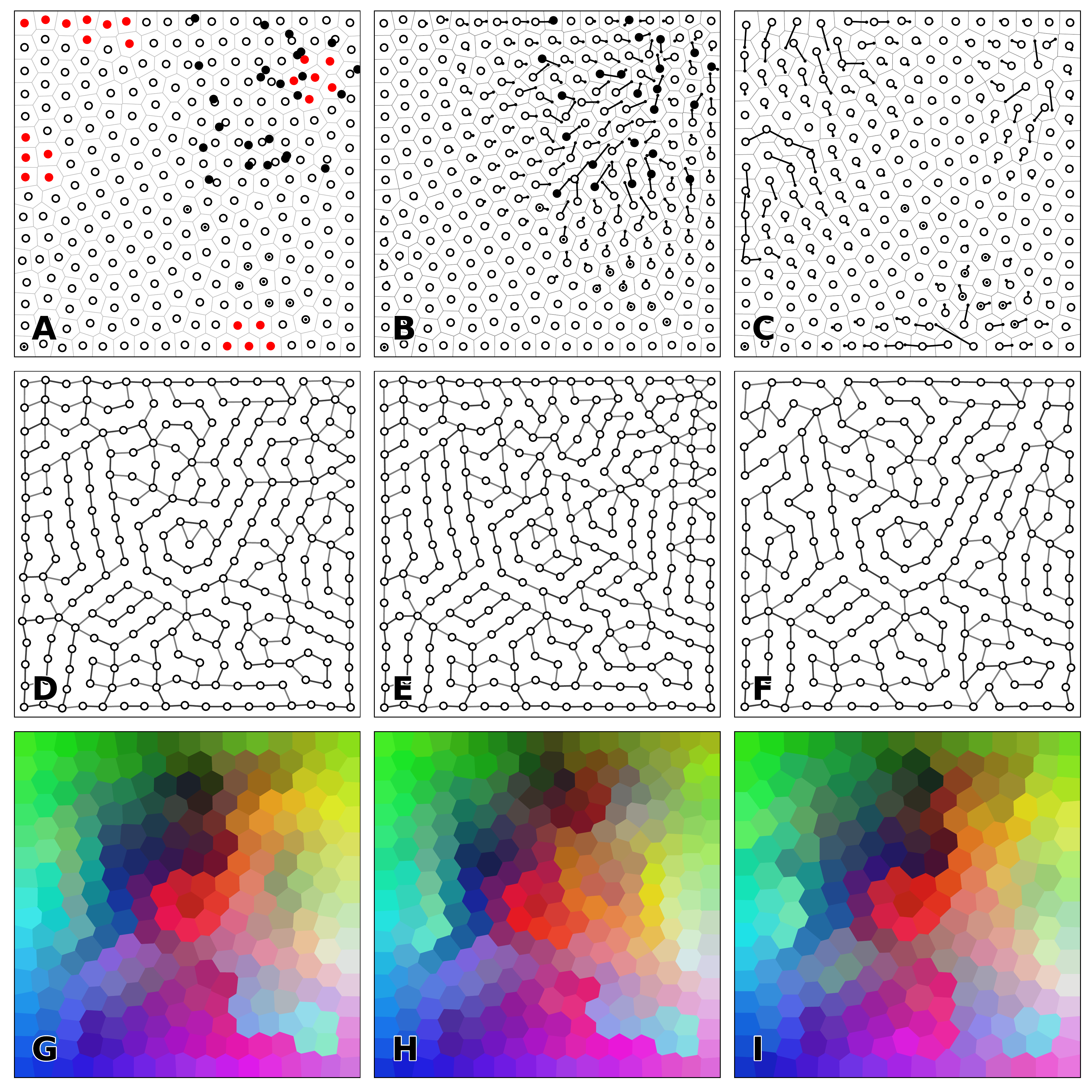}
  \caption{%
  {\bfseries \sffamily Reorganization (results).}
  An initial set of 248 neurons (outlined discs on panel \textbf{A}) has been modified with the addition of 25 neurons (black discs) or the removal of 25 neurons (red discs). Panels \textbf{B} and \textbf{C} show the final position of neurons after 100 iterations of the centroidal Voronoi tesselation. Lines shows individual movement of neurons. Panels \textbf{D}, \textbf{E} and \textbf{F} show the 2-neighbors induced topology for \textbf{A}, \textbf{B} and \textbf{C} respectively. Panels \textbf{G}, \textbf{H} and \textbf{I} show the map
  map codebook for each map in neural space after learnin. Each voronoi cell of a neuron is painted with the color of the related codeword.
  }
  \label{fig:reorganization:results}
\end{figure}

The final and most challenging experiment is to test how the RSOM can cope
with degenerative cases, where either neurons die out (removal) or new units are added to the map (addition). Figure~\ref{fig:reorganization:results}A illustrates an example of a well-formed neural space (black outlined discs), a removal (red disks) and an addition (black dots). For both removal and addition, we applied a LLoyd relaxation scheme to achieve a new quasi-centroidal Voronoi tesselation. Figures \ref{fig:reorganization:results}B and \ref{fig:reorganization:results}C depicts the Voronoi tesselations after $100$ iterations starting from the initial tesselation shown in panel \ref{fig:reorganization:results}A. 

In order to conserve as much as possible the original topology, we used a differentiated procedure depending on if we are dealing with a removal or an addition. In case of removal, only the remaining neurons that were previously connected to a removed unit are allowed to connect to a new unit unconditionally. For the rest of the units, they might reconnect to a nearby unit if this unit is much closer than its closest current neighbour (85\% of the smallest distance to its current neighbours). In case of addition, new units can connect unconditionally to the nearest neighbours while old unit can only connect to the newly added unit if this unit is much closer than its current closest neighbour (85\% of the smallest distance to its current neighbours). This procedure guarantees that the topology is approximately conserved as shown in figures \ref{fig:reorganization:results}D-F. We tested the alternative of recomputing the graph from scratch but the resulting topology is quite different from the original because of micro-displacements of every units following the Lloyd relaxation.

Learning is performed in two steps. First we iterate $25,000$ epochs using the intact map, then we perform removal and addition and learning is iterated for another $5,000$ epochs for all three maps (orginal map, map with added units and map with removed units). The final self-organization is shown in figures \ref{fig:reorganization:results}G-I where we can observe strong similarities in the organization. For example, the central red patch is conserved in all three maps and the overall structure is visually similar. Of course, these results depend on the number of removed or added units (that needs to be relatively small compared to the size of the whole map) and their spatial distribution.
%\gid{the last two paragraphs are not too clear, they seem confusing}

%\gid{TODO Add a few more words here. Maybe make a connection to neuroscience facts about reorganization.} \npr{maybe in the discussion instead} \gid{Agreed} \nrp{Done}.

%Previous studies have shown that after an ablation or addition of units the map undergoes a reorganization of its neural representations. The immediate  consequence of a reorganization is the acquisition of new receptive fields. When an ablation takes place, unaffected neurons try to recover representations that got lost due to neurons death. In addition, on the other hand, the surplus neurons have to get their share of the representations and thus the entire map undergoes a reorganization.

%The induced topology shares similarity with the original topology (before ablation or addition). 

% It is known that this sort of reorganization takes place naturally within brain. More precisely, neurogenesis happens in the subgranular zone of the Dentate Gyrus (DG) of the hippocampus and in the subventricular zone of the lateral ventricle~\cite{Alvarez:2004}. On the other hand, when neural tissue in the cerebral cortex~\cite{Merzenich:1984,Taub:2014} or the spinal cord~\cite{Bareyre:2004,Liu:1958}, the undamaged neurons reorganize their receptive fields and undamaged nerves sprout new connections, respectively, partially or fully restore functionality. Therefore, the study of reorganization after a neural ablation or neurogenesis within a self-organizing map is detrimental. 

\section{Discussion}

We have introduced a variation of the self-organizing map algorithm by considering the random placement of neurons on a two-dimensional manifold, following a blue noise distribution from which various topologies can be derived. We've shown these topologies possess random (but controllable) discontinuities that allow for a more flexible self-organization, especially with high-dimensional data. This has been demonstrated for low-dimensional cases as well as for high-dimensional case such as the classical MNIST dataset \citep{Lecun:1998}. To analyze the results and characterize properties within the maps, we used tools from the field of topological data analysis and random matrix theory that provide extra information when compared to regular quality measures \citep{Polani2002}. More specifically, we computed the persistence diagrams and barcodes for both the regular and randomized self-organizing maps and the input space and we estimated the eigenvalues distributions of the Gram matrices for the activity of both SOMs. Overall, our results show that the proposed algorithm performs equally well as the original SOM and develop well-formed topographic maps. In some cases, RSOM preserves actually better the topological properties of the input space when compared to the original SOM but it is difficult to assert that this is a general property since a theoretical approach would be a hard problem. Another important aspect we highlighted is that RSOM can cope with the addition or the removal of units during learning and preserve, to a large extent, the underlying self-organization. This reorganization capacity allows to have an adaptive architecture where neurons can be added or removed following an arbitrary quality criterion.

This article comes last in a series of three articles where we investigated
the conditions for a more biologically plausible self-organization process. In the first article \citep{Rougier:2017}, we introduced the dynamic SOM (DSOM) and showed how the time-dependent learning rate and neighborhood function variance of regular SOM can be replaced by a time-independent learning process. DSOM is capable of continuous on-line learning and can adapt anytime to a dynamic dataset. In the second article \citep{Detorakis2012, Detorakis2014}, we introduced the dynamic neural field SOM (DNF-SOM) where the winner-take-all competitive stage has been replaced by a regular neural field that aimed at simulating the neural activity of the somatosensory cortex (area 3b). The whole SOM procedure is thus replaced by an actual distributed process without the need of any supervisor to select the BMU. The selection of the BMU as well as the neighborhood function emerge naturally
%It naturally emerges as a Gaussian bump
due to the lateral competition between neurons that ultimately drives the self-organization. The present work is the last part of this sequel and provides the basis for developing biologically plausible self-organizing maps. Taken together, DSOM, DNF-SOM and RSOM provides a biological ground for self-organization where decreasing learning rate, winner-take-all and regular grid are not necessary. Instead, our main hypotheses are the blue noise distribution and the nearest-neighbour connectivity pattern. For the blue noise distribution and given the physical nature of neurons \citep{BlazquezLlorca2014,Lanaro:2020}, we think it makes sense to consider neurons to be at a minimal distance from each others and randomly distributed and to have a nearest-neighbour connectivity as it is known to occur in the cortex \citep{vanPelt2013}. The case of reorganization, where neurons physically migrate (Lloyd relaxation), is probably the most dubious hypothesis but seems to be partially supported by experimental results
\citep{Kaneko2017}. It is also worth to mention that reorganization takes place naturally in the mammal brain. More precisely, neurogenesis happens in the subgranular zone of the dentate gyrus of the hippocampus and in the subventricular zone of the lateral ventricle \citep{Alvarez:2004}. On the other hand, when neural tissue in the cerebral cortex \citep{Merzenich:1984,Taub:2014} or the spinal cord \citep{Bareyre:2004,Liu:1958} are damaged, neurons reorganize their receptive fields and undamaged nerves sprout new connections and restore function (partially or fully). During such event, it has been shown that neurons can physically move.

% \gid{Maybe we could fit these sentences somewhere:
Finally, the analysis we performed (TDA, eigenvalues distributions, distortion, and entropy indicates that both SOM and RSOM perform equally well. For the majority of the measures we used to assess the performance of both algorithms, we observed very similar results. Only in the case of TDA, we identified some differences in the topological features the two algorithms can capture. More precisely, both algorithms generate maps that capture most of the topological features of the input space. RSOM tends to capture slightly better high-dimensional topological features, especially for input spaces with holes (see the experiment on the $2D$ uniform distribution in Section~\ref{sec:2d-holes}). Therefore, we can conclude that the RSOM matches the performance of the SOM.

\section*{Abbreviations}
\begin{description}
    \item[BMU]     Best Matching unit
    \item[DNF-SOM] Dynamic Neural Field-Self-Organizing Map
    \item[DSOM]    Dynamic Self-organizing Map
    \item[KSOM]    Kohonen Self-Organizing Map (Kohonen original proposal)
    \item[KDE]     Kernel Density Estimation
    \item[RSOM]    Randomized Self-Organizing Map
    \item[TDA]     Topological Data Analysis
    \item[SOM]     Self-Organizing Map
\end{description}

\section*{Funding}
This work was partially funded by grant ANR-17-CE24-0036.
% \printbibliography %[heading=bibintoc]

\bibliographystyle{plain}
\bibliography{biblio}

\begin{thebibliography}{10}

\bibitem{Alahakoon:2000}
D.~Alahakoon, S.K. Halgamuge, and B.~Srinivasan.
\newblock Dynamic self-organizing maps with controlled growth for knowledge
  discovery.
\newblock {\em {IEEE} Transactions on Neural Networks}, 11(3):601--614, 5 2000.

\bibitem{Alvarez:2004}
Arturo Alvarez-Buylla and Daniel~A Lim.
\newblock For the long run: maintaining germinal niches in the adult brain.
\newblock {\em Neuron}, 41(5):683--686, 2004.

\bibitem{Astudillo:2014}
C{\'{e}}sar~A. Astudillo and B.~John Oommen.
\newblock Topology-oriented self-organizing maps: a survey.
\newblock {\em Pattern Analysis and Applications}, 17(2):223--248, 3 2014.

\bibitem{Bareyre:2004}
Florence~M Bareyre, Martin Kerschensteiner, Olivier Raineteau, Thomas~C
  Mettenleiter, Oliver Weinmann, and Martin~E Schwab.
\newblock The injured spinal cord spontaneously forms a new intraspinal circuit
  in adult rats.
\newblock {\em Nature neuroscience}, 7(3):269--277, 2004.

\bibitem{Blackmore:1995}
Justine Blackmore and Risto Miikkulainen.
\newblock Visualizing high-dimensional structure with the incremental grid
  growing neural network.
\newblock In {\em Machine Learning Proceedings 1995}, pages 55--63. Elsevier,
  1995.

\bibitem{BlazquezLlorca2014}
Lidia Blazquez-Llorca, Alan Woodruff, Melis Inan, Stewart~A. Anderson, Rafael
  Yuste, Javier DeFelipe, and Angel Merchan-Perez.
\newblock Spatial distribution of neurons innervated by chandelier cells.
\newblock {\em Brain Structure and Function}, 220(5):2817--2834, July 2014.

\bibitem{Bridson:2007}
Robert Bridson.
\newblock Fast poisson disk sampling in arbitrary dimensions.
\newblock In {\em {ACM} {SIGGRAPH} 2007 sketches on - {SIGGRAPH}
  {\textquotesingle}07}. {ACM} Press, 2007.

\bibitem{Burguillo:2013}
Juan~C. Burguillo.
\newblock Using self-organizing maps with complex network topologies and
  coalitions for time series prediction.
\newblock {\em Soft Computing}, 18(4):695--705, 11 2013.

\bibitem{Carlsson:2009}
Gunnar Carlsson.
\newblock Topology and data.
\newblock {\em Bulletin of the American Mathematical Society}, 46(2):255--308,
  2009.

\bibitem{Chazal:2017}
Fr{\'e}d{\'e}ric Chazal and Bertrand Michel.
\newblock An introduction to topological data analysis: fundamental and
  practical aspects for data scientists.
\newblock {\em arXiv preprint arXiv:1710.04019}, 2017.

\bibitem{Demartines:1992}
Pierre Demartines.
\newblock Organization measures and representations of kohonen maps.
\newblock In {\em First IFIP Working Group}, volume~10. Citeseer, 1992.

\bibitem{Detorakis:2019}
Georgios Detorakis, Travis Bartley, and Emre Neftci.
\newblock Contrastive hebbian learning with random feedback weights.
\newblock {\em Neural Networks}, 114:1--14, 2019.

\bibitem{Detorakis2012}
Georgios~Is. Detorakis and Nicolas~P. Rougier.
\newblock A neural field model of the somatosensory cortex: Formation,
  maintenance and reorganization of ordered topographic maps.
\newblock {\em {PLoS} {ONE}}, 7(7):e40257, July 2012.

\bibitem{Detorakis2014}
Georgios~Is. Detorakis and Nicolas~P. Rougier.
\newblock Structure of receptive fields in a computational model of area 3b of
  primary sensory cortex.
\newblock {\em Frontiers in Computational Neuroscience}, 8, July 2014.

\bibitem{Edelsbrunner:2008}
Herbert Edelsbrunner and John Harer.
\newblock Persistent homology-a survey.
\newblock {\em Contemporary mathematics}, 453:257--282, 2008.

\bibitem{Eiben:2003}
A.E. Eiben and J.E. Smith.
\newblock {\em Introduction to Evolutionary Computing}.
\newblock Springer Verlag, 2003.

\bibitem{Fritzke:1994}
Bernd Fritzke.
\newblock A growing neural gas network learns topologies.
\newblock In {\em Proceedings of the 7th International Conference on Neural
  Information Processing Systems}, NIPS'94, pages 625--632, Cambridge, MA, USA,
  1994. MIT Press.

\bibitem{Ghrist:2008}
Robert Ghrist.
\newblock Barcodes: the persistent topology of data.
\newblock {\em Bulletin of the American Mathematical Society}, 45(1):61--75,
  2008.

\bibitem{HerculanoHouzel:2013}
Suzana Herculano-Houzel, Charles Watson, and George Paxinos.
\newblock Distribution of neurons in functional areas of the mouse cerebral
  cortex reveals quantitatively different cortical zones.
\newblock {\em Frontiers in Neuroanatomy}, 7, 2013.

\bibitem{Hunter:2007}
J.~D. Hunter.
\newblock Matplotlib: A 2d graphics environment.
\newblock {\em Computing In Science \& Engineering}, 9(3):90--95, 2007.

\bibitem{Jiang:2009}
Fei Jiang, Hugues Berry, and Marc Schoenauer.
\newblock The impact of network topology on self-organizing maps.
\newblock In {\em Proceedings of the first {ACM}/{SIGEVO} Summit on Genetic and
  Evolutionary Computation - {GEC} {\textquotesingle}09}. {ACM} Press, 2009.

\bibitem{Jones:2001}
Eric Jones, Travis Oliphant, and Pearu Peterson.
\newblock {SciPy}: Open source scientific tools for {Python}, 2001.

\bibitem{Kaneko2017}
Naoko Kaneko, Masato Sawada, and Kazunobu Sawamoto.
\newblock Mechanisms of neuronal migration in the adult brain.
\newblock {\em Journal of Neurochemistry}, 141(6):835--847, April 2017.

\bibitem{Kaski:1998}
Samuel Kaski, Jari Kangas, and Teuvo Kohonen.
\newblock Bibliography of self-organizing map (som) papers: 1981-1997.
\newblock {\em Neural Computing Surveys}, 1, 1998.

\bibitem{Kohonen:1982}
Teuvo Kohonen.
\newblock Self-organized formation of topologically correct feature maps.
\newblock {\em Biological Cybernetics}, 43(1):59--69, 1982.

\bibitem{Kohonen:2001}
Teuvo Kohonen.
\newblock {\em Self-Organizing Maps}, volume~30 of {\em Springer Series in
  Information Sciences}.
\newblock Springer-Verlag, Berlin, Germany, 3 edition, 2001.

\bibitem{Lagae:2008}
Ares Lagae and Philip Dutr{\'{e}}.
\newblock A comparison of methods for generating poisson disk distributions.
\newblock {\em Computer Graphics Forum}, 27(1):114--129, 3 2008.

\bibitem{Lanaro:2020}
Matteo~Paolo Lanaro, H{\'e}l{\`e}ne Perrier, David Coeurjolly, Victor
  Ostromoukhov, and Alessandro Rizzi.
\newblock Blue-noise sampling for human retinal cone spatial distribution
  modeling.
\newblock {\em Journal of Physics Communications}, 4(3):035013, 2020.

\bibitem{Lecun:1998}
Yann LeCun, L{\'e}on Bottou, Yoshua Bengio, and Patrick Haffner.
\newblock Gradient-based learning applied to document recognition.
\newblock {\em Proceedings of the IEEE}, 86(11):2278--2324, 1998.

\bibitem{Liu:1958}
Chan-Nao Liu and WW~Chambers.
\newblock Intraspinal sprouting of dorsal root axons: Development of new
  collaterals and preterminals following partial denervation of the spinal cord
  in the cat.
\newblock {\em AMA Archives of Neurology \& Psychiatry}, 79(1):46--61, 1958.

\bibitem{Lloyd:1982}
S.~Lloyd.
\newblock Least squares quantization in {PCM}.
\newblock {\em {IEEE} Transactions on Information Theory}, 28(2):129--137, 3
  1982.

\bibitem{Maria:2014}
Cl{\'e}ment Maria, Jean-Daniel Boissonnat, Marc Glisse, and Mariette Yvinec.
\newblock The gudhi library: Simplicial complexes and persistent homology.
\newblock In {\em International Congress on Mathematical Software}, pages
  167--174. Springer, 2014.

\bibitem{Mcinnes:2018}
Leland McInnes, John Healy, and James Melville.
\newblock Umap: Uniform manifold approximation and projection for dimension
  reduction.
\newblock {\em arXiv preprint arXiv:1802.03426}, 2018.

\bibitem{Merzenich:1984}
Michael~M Merzenich, Randall~J Nelson, Michael~P Stryker, Max~S Cynader, Axel
  Schoppmann, and John~M Zook.
\newblock Somatosensory cortical map changes following digit amputation in
  adult monkeys.
\newblock {\em Journal of comparative Neurology}, 224(4):591--605, 1984.

\bibitem{Oja:2003}
Merja Oja, Samuel Kaski, and Teuvo Kohonen.
\newblock Bibliography of self-organizing map (som) papers: 1998-2001 addendum.
\newblock {\em Neural Computing Surveys}, 3, 2003.

\bibitem{Parzen:1962}
Emanuel Parzen.
\newblock On estimation of a probability density function and mode.
\newblock {\em The annals of mathematical statistics}, 33(3):1065--1076, 1962.

\bibitem{Pedregosa:2011}
Fabian Pedregosa, Ga{\"e}l Varoquaux, Alexandre Gramfort, Vincent Michel,
  Bertrand Thirion, Olivier Grisel, Mathieu Blondel, Peter Prettenhofer, Ron
  Weiss, Vincent Dubourg, et~al.
\newblock Scikit-learn: Machine learning in python.
\newblock {\em the Journal of machine Learning research}, 12:2825--2830, 2011.

\bibitem{Polani2002}
Daniel Polani.
\newblock Measures for the organization of self-organizing maps.
\newblock In {\em Self-Organizing Neural Networks}, pages 13--44.
  Physica-Verlag {HD}, 2002.

\bibitem{Polla:2009}
Matti Pöllä, Timo Honkela, and Teuvo Kohonen.
\newblock Bibliography of self-organizing map (som) papers: 2002-2005 addendum.
\newblock {\em Helsinki University of Technology}, 2009.

\bibitem{Rougier:2017}
Nicolas~P. Rougier.
\newblock {[Re] Weighted Voronoi Stippling}.
\newblock {\em ReScience}, 3(1), 2017.

\bibitem{rougier:2011}
Nicolas~P. Rougier and Yann Boniface.
\newblock Dynamic self-organising map.
\newblock {\em Neurocomputing}, 74(11):1840--1847, 2011.

\bibitem{rynkiewicz:2008}
Joseph Rynkiewicz.
\newblock Self organizing map algorithm and distortion measure, 2008.

\bibitem{Taub:2014}
Edward Taub, Gitendra Uswatte, and Victor~W Mark.
\newblock The functional significance of cortical reorganization and the
  parallel development of ci therapy.
\newblock {\em Frontiers in Human Neuroscience}, 8:396, 2014.

\bibitem{Walt:2011}
Stéfan van~der Walt, S~Chris Colbert, and Gaël Varoquaux.
\newblock The {NumPy} array: A structure for efficient numerical computation.
\newblock {\em Computing in Science {\&} Engineering}, 13(2):22--30, 3 2011.

\bibitem{vanPelt2013}
Jaap van Pelt and Arjen van Ooyen.
\newblock Estimating neuronal connectivity from axonal and dendritic density
  fields.
\newblock {\em Frontiers in Computational Neuroscience}, 7, 2013.

\bibitem{Villmann:1999}
Thomas Villmann.
\newblock Topology preservation in self-organizing maps.
\newblock In {\em Kohonen Maps}, pages 279--292. Elsevier, 1999.

\bibitem{Zhou:2012}
Yahan Zhou, Haibin Huang, Li-Yi Wei, and Rui Wang.
\newblock Point sampling with general noise spectrum.
\newblock {\em {ACM} Transactions on Graphics}, 31(4):1--11, 7 2012.

\bibitem{Zomorodian:2005}
Afra Zomorodian and Gunnar Carlsson.
\newblock Computing persistent homology.
\newblock {\em Discrete \& Computational Geometry}, 33(2):249--274, 2005.

\end{thebibliography}

\newpage
\appendix
\section{Supplementary material}

Randomized Self Organizing Map \\ 
Nicolas P. Rougier$^{1,2,3}$ and Georgios Is. Detorakis$^4$ \\
$^1$ Inria Bordeaux Sud-Ouest \\
$^2$ Institut des Maladies Neurodégénératives, Université  de Bordeaux, CNRS UMR 5293 \\
$^3$ LaBRI, Université de Bordeaux, Institut Polytechnique de Bordeaux, CNRS UMR 5800 \\
$^4$ adNomus Inc., San Jose, CA, USA \\
\medskip
% Self-organizing maps usually rely on a predetermined topology of the neural space (the map), which is either a rectangular or a hexagonal Cartesian grid. When the intrinsic dimension of  the input space is much higher than the allowed dimension of the neural space, then the self-organizing map can be ill-formed. To overcome this problem in  high dimensional input spaces. We propose a variation of the self organizing map algorithm, where we consider random placement of neurons on a high-dimensional manifold. The positions of the neural units are drawn from a blue noise distribution from which various topologies can be derived. These topologies possess  random but controllable discontinuities that allow for a flexible self-organization, especially with high-dimensional data. The proposed algorithm has been tested on one-, two- and three-dimensions tasks as well as on MNIST handwritten digits dataset. Furthermore, we investigate the reorganization of the self-organizing maps when we either remove or add neurons to the map. To analyze the results we use spectral analysis and topological data analysis tools. 
%

\textbf{Abstract.} We propose a variation of the self organizing map algorithm by considering the random placement of neurons on a two-dimensional manifold, following a blue noise distribution from which various topologies can be derived. These topologies possess random (but controllable) discontinuities that allow for a more flexible self-organization, especially with high-dimensional data. The proposed algorithm is tested on one-, two- and three-dimensions tasks as well as on the MNIST handwritten digits dataset and validated using spectral analysis and topological data analysis tools. We also demonstrate the ability of the randomized self-organizing map to gracefully reorganize itself in case of neural lesion and/or neurogenesis.\par

% \textbf{Keywords.} Self Organization, Neural Networks, Vector Quantization, Voronoi Tesselation, Neural Map Topology

\bigskip

\subsection{One-dimensional uniform dataset}

\begin{figure}
  \includegraphics[width=\columnwidth]{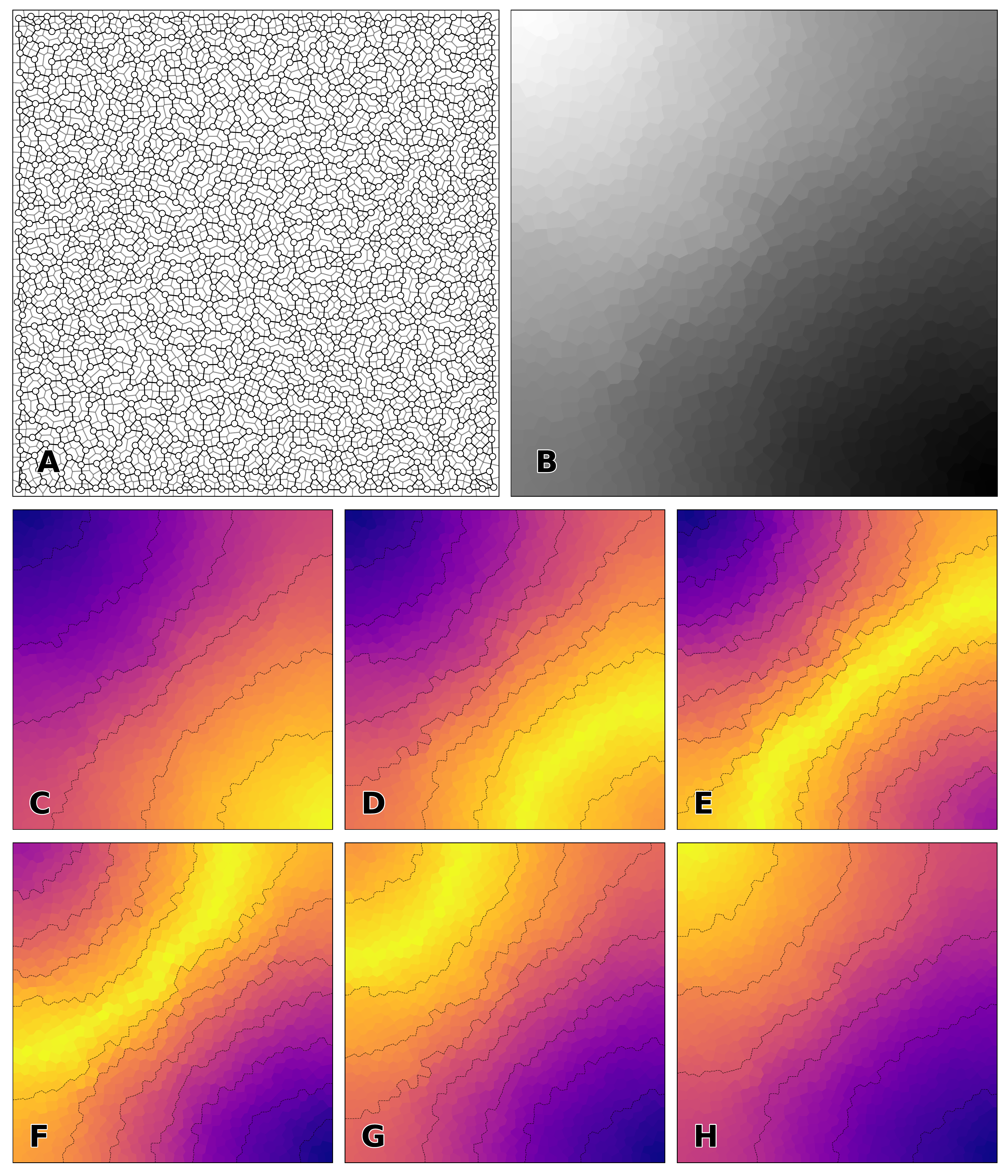}
  \vspace{2mm}
  \centering
  \includegraphics[width=.975\columnwidth]{colormap.pdf}
  \caption{%
  {\bfseries \sffamily One dimensional uniform dataset with holes (results)}
  Randomized SOM made of $1024$ neurons with a $3$-nearest neighbors induced topology. Model has been trained for $25,000$ epochs on one-dimensional points drawn from a uniform distribution on the unit segment. \textbf{A} Map topology in neural space. \textbf{B} Map topology in data space. \textbf{C to H} Receptive field of the map for six samples.
  }
  \label{fig:1D-uniform:results}
\end{figure}

\subsection{Two dimensional uniform dataset}

\begin{figure}
  \includegraphics[width=\columnwidth]{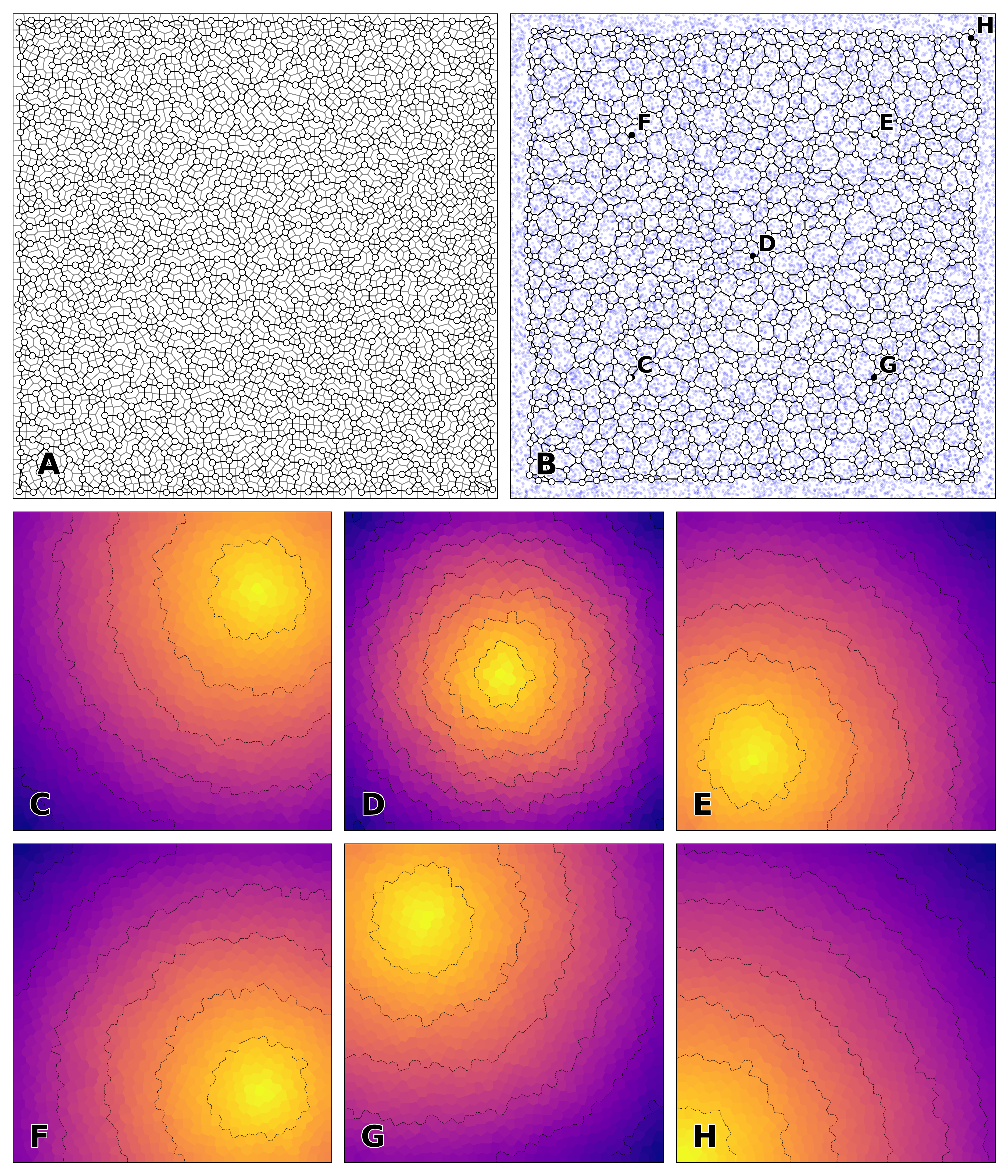}
  \vspace{2mm}
  \centering
  \includegraphics[width=.975\columnwidth]{colormap.pdf}
  \caption{%
  {\bfseries \sffamily Two dimensional uniform dataset (results)}
  Randomized SOM made of $1024$ neurons with a $2$-nearest neighbors induced topology. Model has been trained for $25,000$ epochs on two-dimensional points drawn from a uniform distribution on the unit square. \textbf{A} Map topology in neural space. \textbf{B} Map topology in data space. \textbf{C to H} Receptive field of the map for six samples.
  }
  \label{fig:2D-uniform:results}
\end{figure}

\subsection{Two-dimensional ring dataset }

\begin{figure}
  \includegraphics[width=\columnwidth]{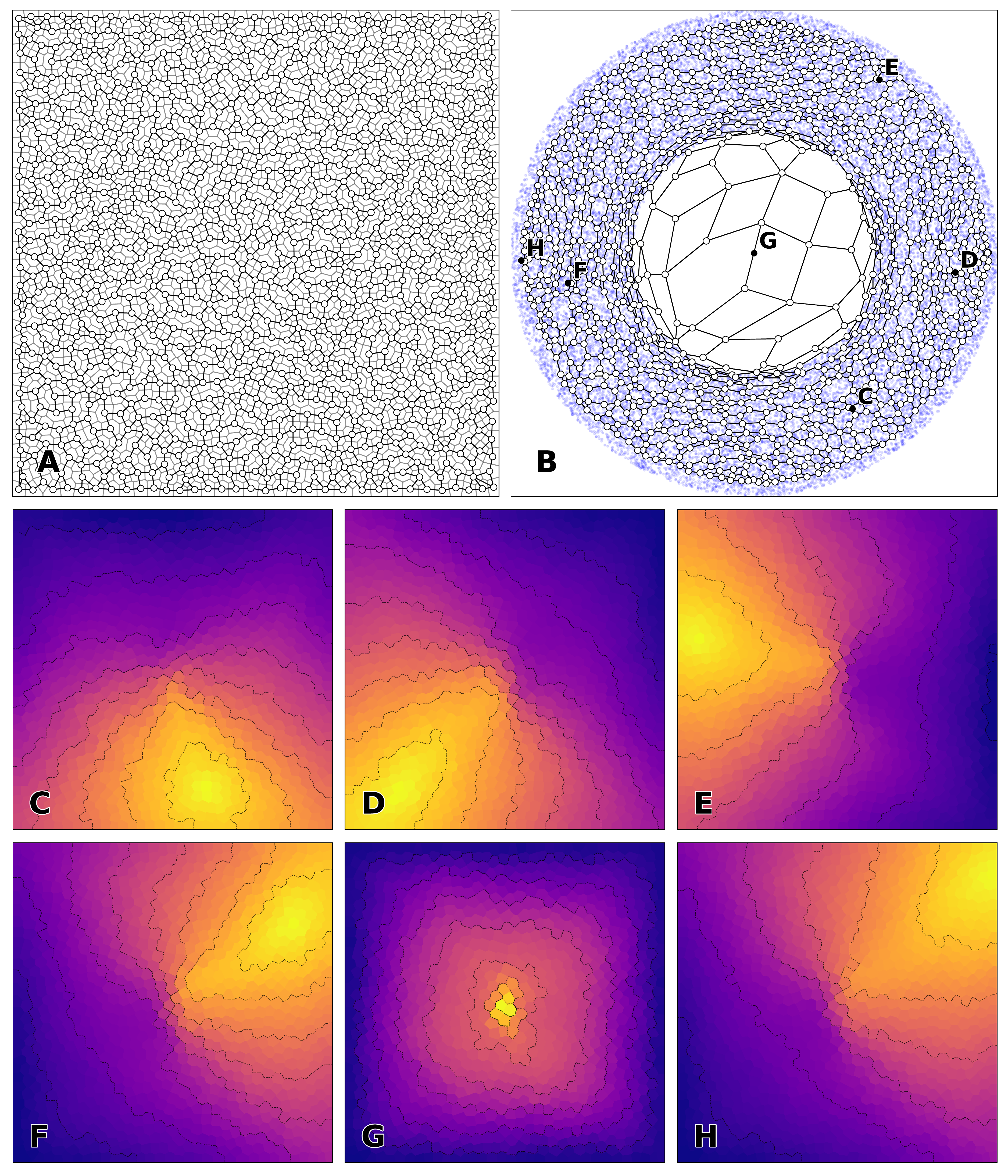}
  \vspace{2mm}
  \centering
  \includegraphics[width=.975\columnwidth]{colormap.pdf}
  \caption{%
  {\bfseries \sffamily Two dimensional ring dataset (results)}
  Randomized SOM made of $1024$ neurons with a $3$-nearest neighbors induced topology. Model has been trained for $25,000$ epochs on two-dimensional points drawn from a ring distribution on the unit square. \textbf{A} Map topology in neural space. \textbf{B} Map topology in data space. \textbf{C to H} Normalized distance map for six samples. Normalization has been performed for each sample in order to enhance contrast but this prevents comparison between maps.
  }
  \label{fig:2D-ring:results}
\end{figure}

\subsection{Oriented Gaussians dataset}

\begin{figure}
  \includegraphics[width=\columnwidth]{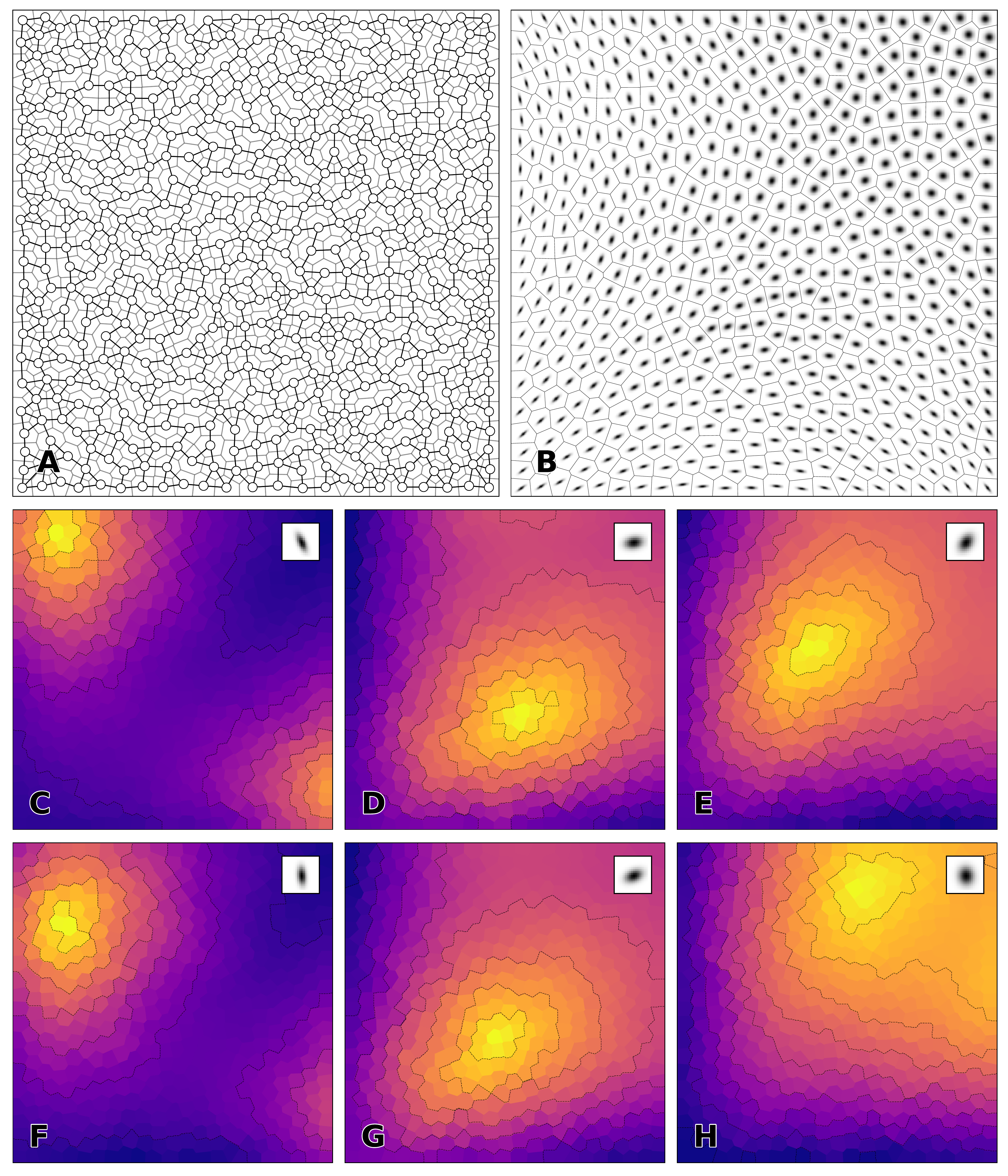}
  \vspace{2mm}
  \includegraphics[width=\columnwidth]{colormap.pdf}
  \caption{%
  {\bfseries \sffamily Oriented Gaussians dataset (results)}
  Randomized SOM made of $1024$ neurons with a $2$-nearest neighbors induced topology. Model has been trained for $25,000$ epochs on oriented Gaussian datasets. \textbf{A} Map topology in neural space. \textbf{B} Map topology in data space. \textbf{C to H} Receptive field of the map for six samples.
  }
  \label{fig:gaussians:results}
\end{figure}

\subsection{Influence of the topology}

\begin{figure}
  \includegraphics[width=\columnwidth]{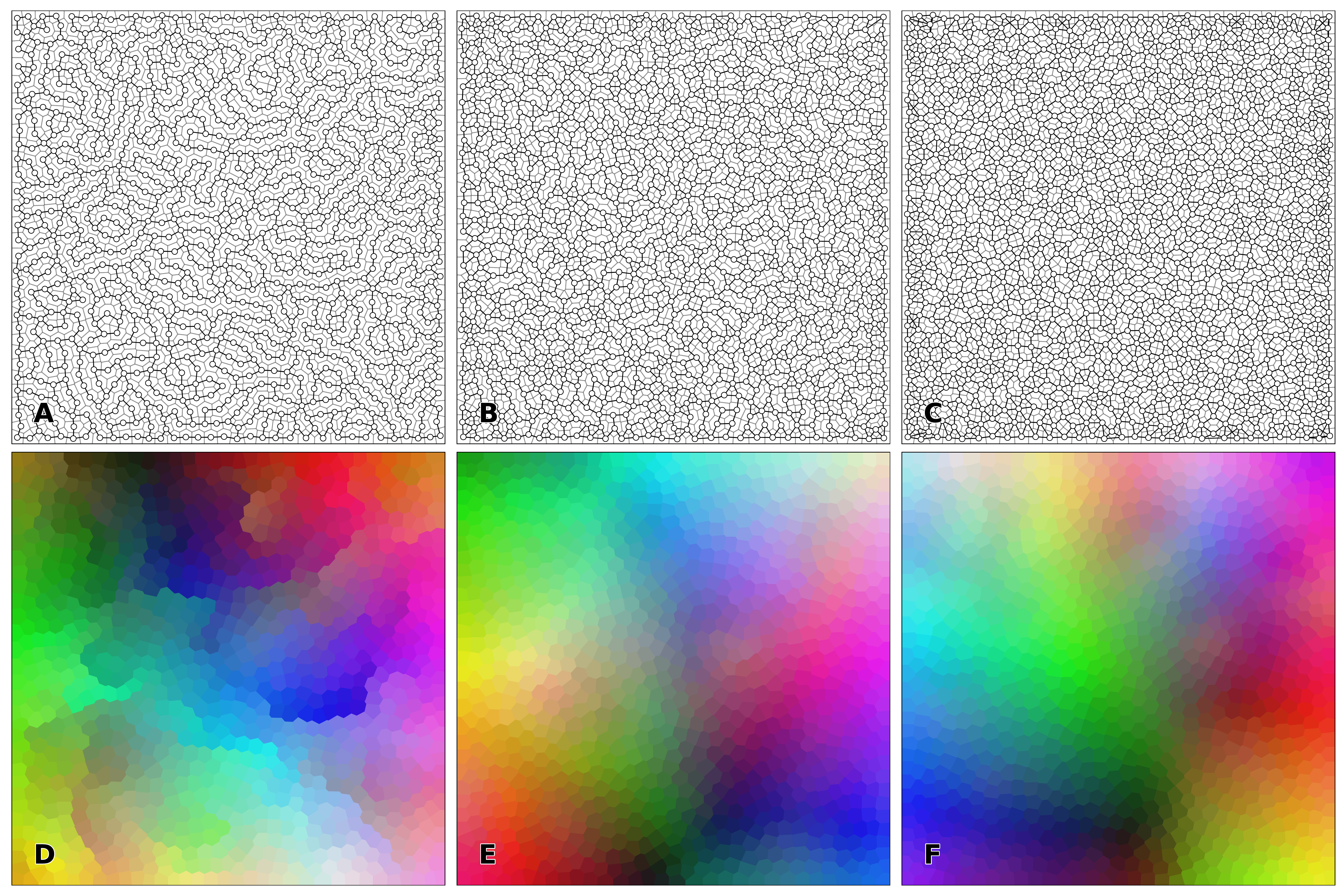}
  \caption{%
  \textbf{Influence of topology on the self organization.}
  The same initial set of 1024 neurons has been equiped with 2-nearest neighbors, 3 nearest neighbors and 4-nearest neighbors induced topology (panels \textbf{A}, \textbf{B} and \textbf{C} respectively) and trained on 25,000 random RGB colors. This lead to qualitatively different self-organization as shown on panels \textbf{D}, \textbf{E} and \textbf{F} respectively, with major discontinuities in the 2-nearest neighbors case. ).
  }
  \label{fig:topology-influence}

\end{figure}

\subsection{Eigenvalues distribution}
\label{sec:dist}

One way to investigate if there is any significant difference between the regular and random SOMs is to compare their neural responses to the same random stimuli. Therefore, we measure the neural activity and build a covariance matrix out of it. Then, we compute the eigenvalues of the covariance matrix (or Gram matrix) and we estimate a probability distribution. Thus, we can compare the eigenvalues distributions of the two maps and compare them to each other. If the distributions are close enough in the sense of Wasserstein distance then the two SOMs are similar in terms of neural activation.  A Gram matrix is an $n \times n$ matrix given by where $n$ is the number of neurons of the map and ${\bf Y} \in \mathbb{R}^{n \times m}$ is a matrix for which each column is the activation of all $n$ neurons to a random stimulus.

From a computational point of view we construct the matrix ${\bf Y}$ by applying a set of stimuli to the self-organized map and computing the activity of each neuron within the map. This implies that ${\bf Y} \in \mathbb{R}^{m \times n}$, where $m=1024$ (the number of neurons) and $n={2, 3}$ (two- or three-dimensional input samples). Then we compute the covariance or Gram matrix as ${\bf M} = {\bf Y}{\bf Y}^T \in \mathbb{R}^{n \times n}$, where $n$ is the number of neurons. Then we compute the eigenvalues and obtain their distribution by sampling the activity of neurons of each experiment for $200$ different initial conditions using $50$ input sample each time. At the end of sampling we get an \emph{ensemble} of $200$ Gram matrices and finally we estimate the probability density of the eigenvalues on each \emph{ensemble} by applying a Kernel Density Estimation method~\citep{Parzen:1962} (KDE) with a Gaussian kernel and bandwidth $h=0.4$. This allows us to quantify any differences on the distributions of the regular and randomized SOMs by calculating the Earth-Mover or Wasserstein-1 distance over the two distributions (regular ($P$) and random SOM ($Q$)). The Wasserstein distance is computed as $W(P, Q) = \inf_{\gamma \in \Pi(P, Q)}\{\mathbb{E}_{(x, y) \sim \gamma}\Big[||x - y||\Big]\}$, where $\Pi(P, Q)$ denotes the set of all joint distributions $\gamma (x, y)$, whose marginals are $P$ and $Q$, respectively. Intuitively, $\gamma (x,y)$ indicates  how  much ``mass'' must be transported from $x$ to $y$ to transform the distribution $P$ into the distribution $Q$. 

\begin{figure}
  \includegraphics[width=\columnwidth]{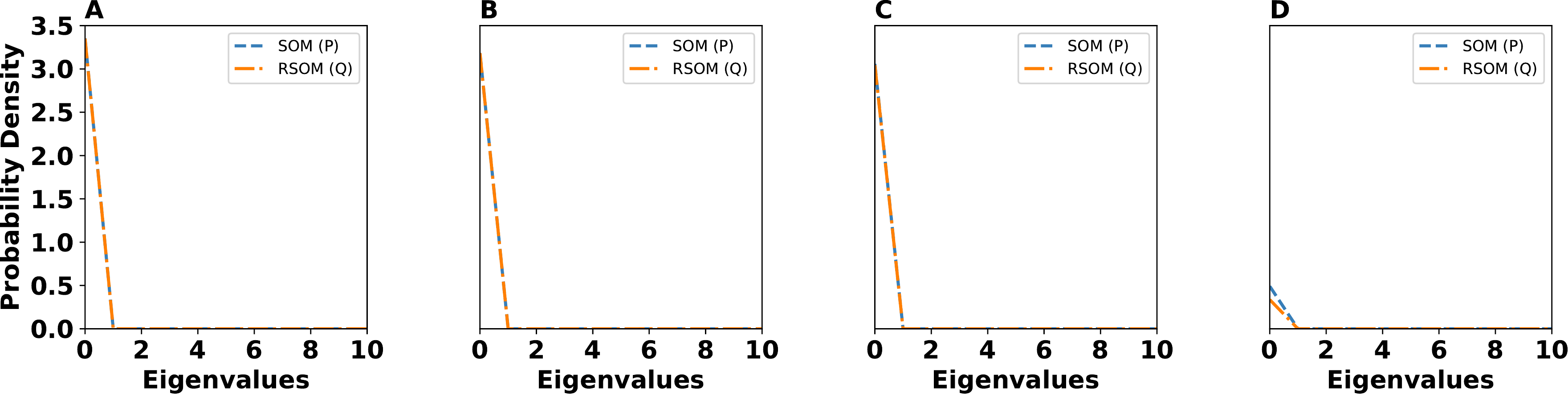}
  \caption{Eigenvalues distribution for \textbf{A} 2D Ring dataset \textbf{B} 2D uniform dataset with holes \textbf{C} 3D uniform dataset and \textbf{D} MNIST Dataset
  }%
  \label{fig:eigenvalues}
 \end{figure}

The distributions of the eigenvalues of the RSOM and the regular SOM are shown on figure~\ref{fig:eigenvalues}. We can conclude that the two distributions are alike and do not suggest any significant difference between the two maps in terms of neural activity. This implies that the RSOM and the regular SOM have similar statistics of their neural activities. This means that the loss of information and the \emph{stretch} to the input data from both RSOM and regular SOM are pretty close and the underlying  topology of the two maps do not really affect the neural activity. This is also confirmed
by measuring the Wasserstein distance between the two distributions. The blue curve shows the regular SOM or distribution $P$ and the black curve the RSOM or distribution $Q$. The Wasserstein distance between the two distributions $P$ and $Q$ indicates that the two distributions are nearly identical on all datasets. The Wasserstein distances in Table~\ref{table:distances}
confirm that the eigenvalues distributions of SOM and RSOM are almost identical indicating that both maps retain the
same amount of information after learning the representations of input spaces.

\begin{table}[!ht]
  \begin{center}
    \begin{tabular}{ll}
        \textbf{Experiment} & \textbf{Wasserstein Distance} \\
        \hline
        $2$D ring dataset               & $0.0000323$\\
        $2$D uniform dataset with holes & $0.0000207$  \\
        $3$D uniform dataset            & $0.0001583$ \\
        MNIST dataset                   & $0.0015$ \\
    \end{tabular}
      \caption{\textbf{Wasserstein distances of eigenvalues distributions.} We report here the Wasserstein 
      distances between eigenvalues distributions of SOM and RSOM for each of the four major experiments we
      ran. The results indicate that the distributions are close pointing out that the SOM and RSOM capture
      a similar level of information during training. For more information regarding how we computed the 
      eigenvalues distributions and the Wasserstein distance please see Section~\ref{sec:dist}.}
      \label{table:distances}
  \end{center}
\end{table}

\subsection{Distortion and entropy measures}

\begin{figure}
  \includegraphics[width=\columnwidth]{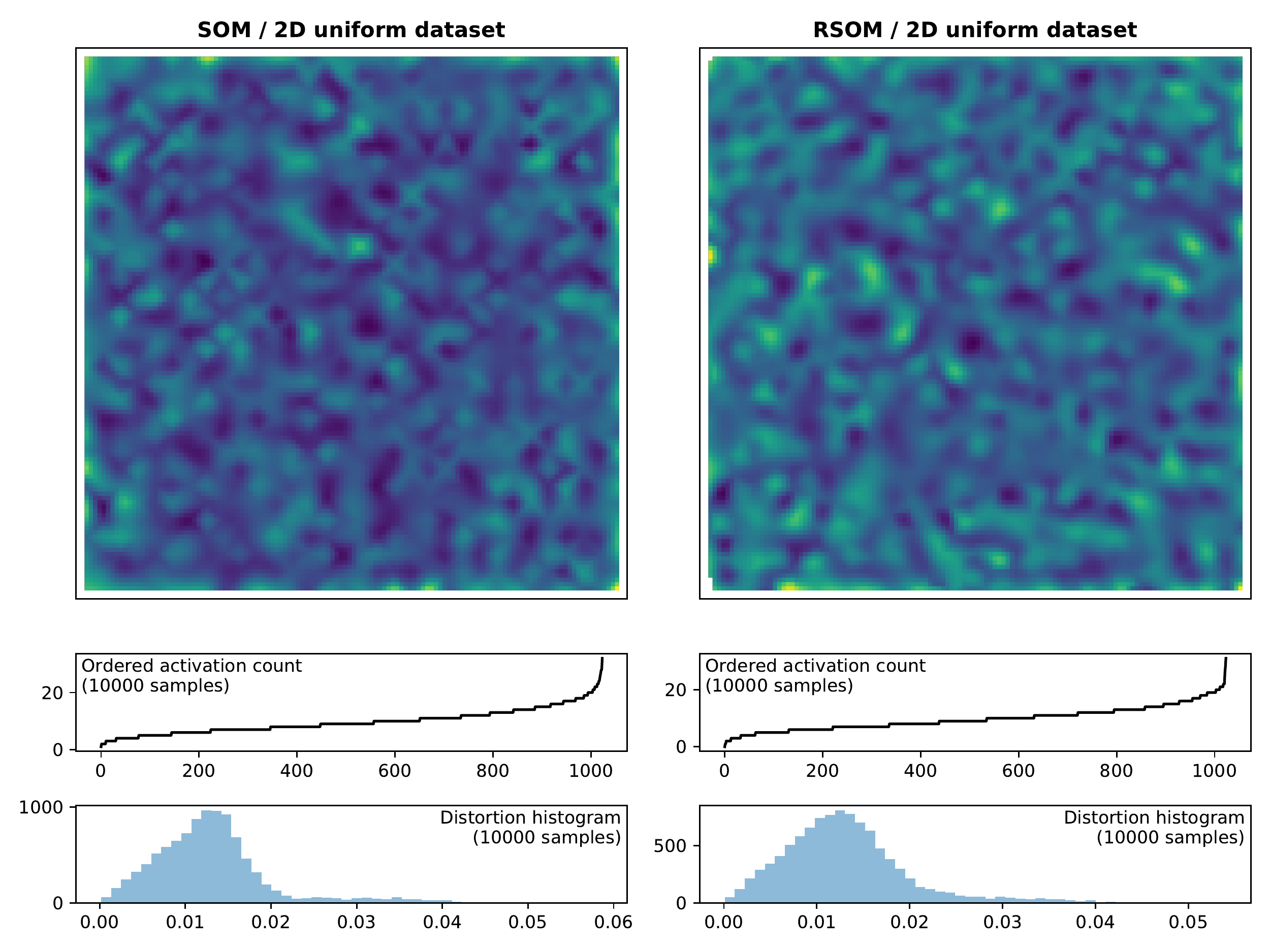}
  \caption{\textbf{Two dimensional uniform dataset (measures)}. Measure of distortion and mean activation over 10,000 samples.}%
\end{figure}

\begin{figure}
  \includegraphics[width=\columnwidth]{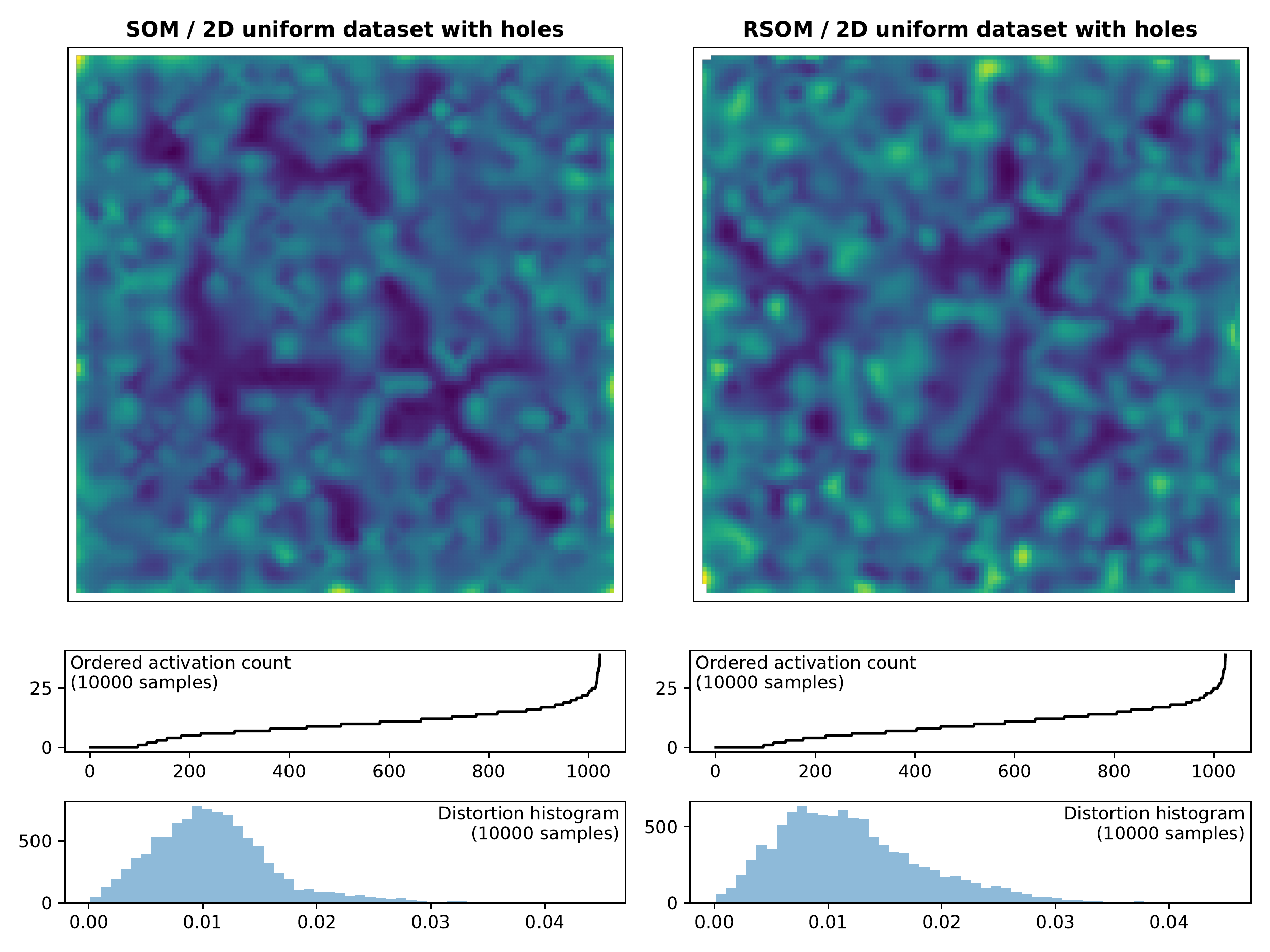}
  \caption{\textbf{Two dimensional uniform dataset with holes (measures)}Measure of distortion and mean activation over 10,000 samples.}%
\end{figure}

\begin{figure}
  \includegraphics[width=\columnwidth]{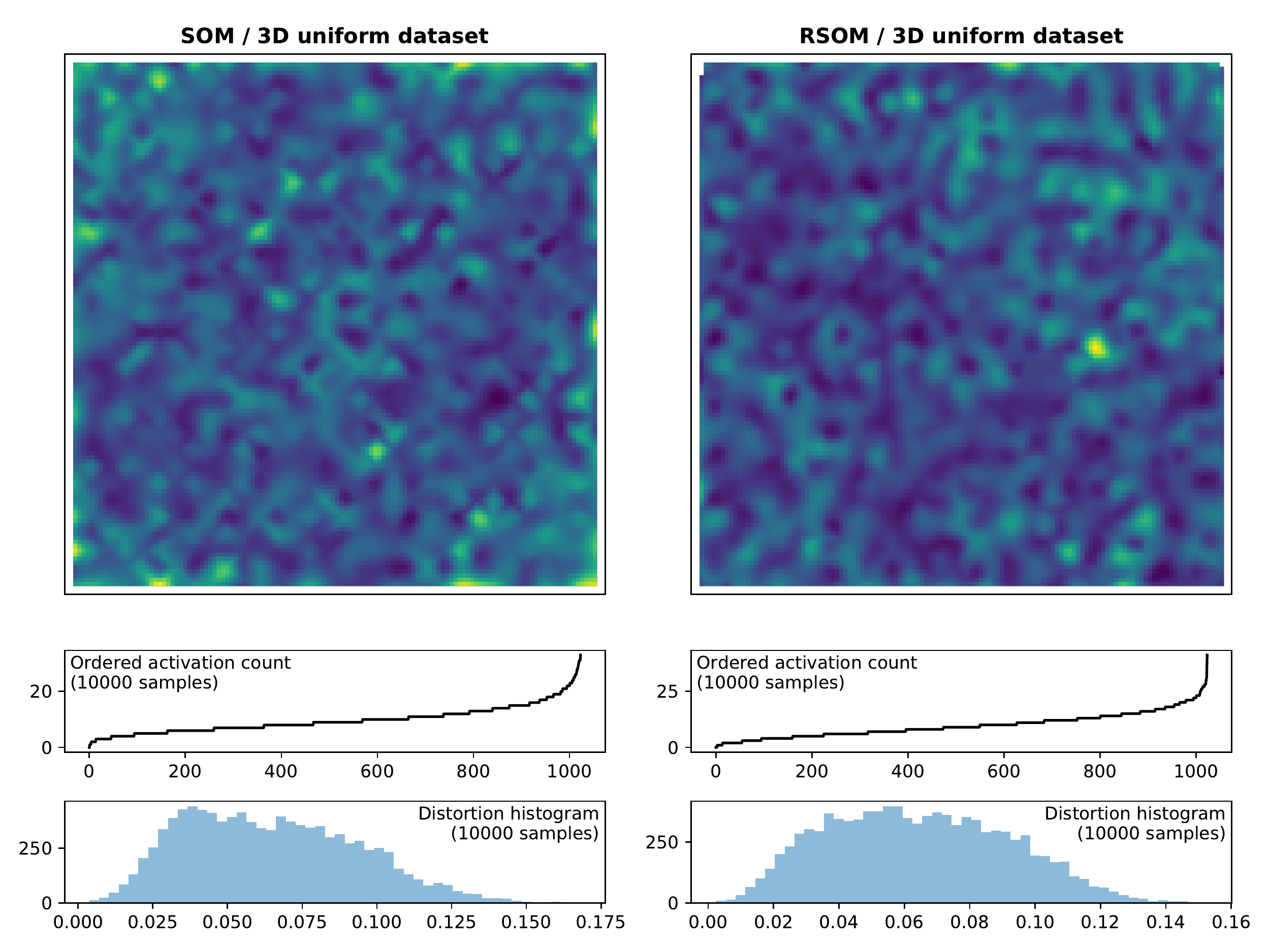}
  \caption{\textbf{Three dimensional uniform dataset (measures)}. Measure of distortion and mean activation over 10,000 samples.}%
\end{figure}

\begin{figure}
  \includegraphics[width=\columnwidth]{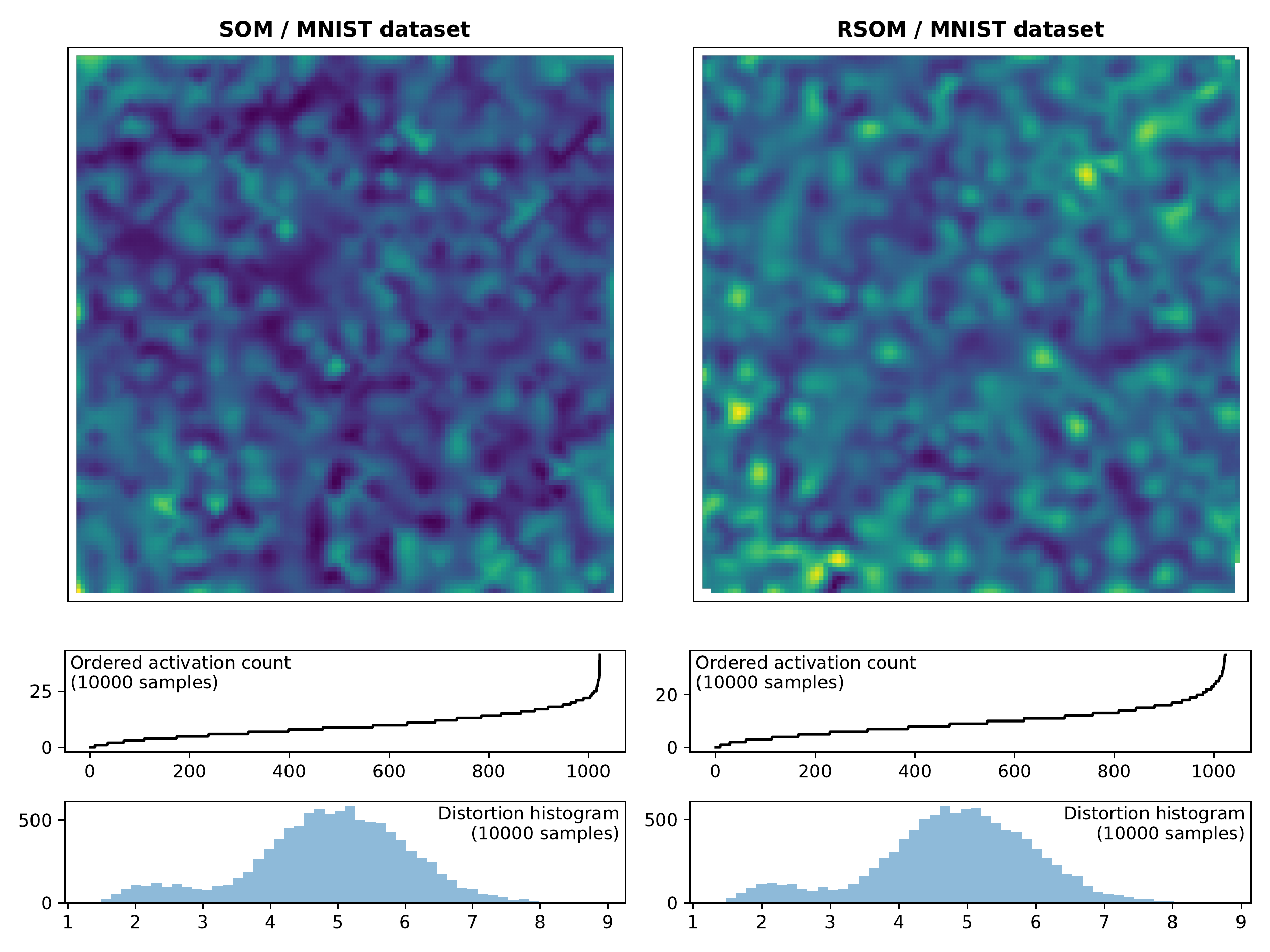}
  \caption{\textbf{MNIST dataset (measures)}. Measure of distortion and mean activation over 10,000 samples.}%
\end{figure}

\end{document}